\pdfoutput=1
\documentclass{article}
\usepackage{arxiv}

\usepackage[utf8]{inputenc}
\usepackage[T1]{fontenc}
\usepackage{float}
\usepackage[hidelinks]{hyperref}
\usepackage{url}
\usepackage{booktabs}
\usepackage{tabularx}
\usepackage{amsmath,amsfonts,amssymb}
\usepackage{microtype}
\usepackage{graphicx}
\usepackage[super,numbers,sort&compress]{natbib}

\usepackage{xcolor}

\newcommand{\AlphaMed}{0.91}

\newcommand{\AlphaWithin}{0.51}
\newcommand{\AlphaWithinC}{0.34}
\newcommand{\AsrFlipMax}{37}
\newcommand{\AsrFlipMed}{5}

\newcommand{\AucBelowSeventy}{39}

\newcommand{\AucDigitsMed}{0.78}

\newcommand{\AucDirectMed}{0.78}

\newcommand{\AucInterviewer}{0.59}
\newcommand{\AucItemSumMed}{0.78}

\newcommand{\AucLettersMed}{0.71}

\newcommand{\AucLettersRevMed}{0.66}

\newcommand{\AucMax}{0.88}
\newcommand{\AucMaxC}{0.91}
\newcommand{\AucMed}{0.75}
\newcommand{\AucMedC}{0.78}
\newcommand{\AucMin}{0.34}
\newcommand{\AucMinC}{0.34}

\newcommand{\BatchFlips}{1}
\newcommand{\BatchMax}{0.68}
\newcommand{\BatchMed}{0.003}
\newcommand{\BatchRows}{500}

\newcommand{\BootCapHi}{21}

\newcommand{\BootCapLo}{15}

\newcommand{\BootPairHi}{42}

\newcommand{\BootPairLo}{39}

\newcommand{\CarryNonePrev}{51}

\newcommand{\CarryOffPrev}{23}

\newcommand{\CarryPrevThPrev}{37}
\newcommand{\CarryPrevThSens}{72}

\newcommand{\CarryTruePrev}{34}

\newcommand{\ContAll}{91}

\newcommand{\DTrue}{0.20}

\newcommand{\DefensAucMax}{0.87}

\newcommand{\DefensAucMin}{0.61}

\newcommand{\DiscKTen}{94}

\newcommand{\DiscKTwo}{40}

\newcommand{\DiscStrict}{53}

\newcommand{\ECDefensPair}{21}
\newcommand{\ECDefensPairC}{21}

\newcommand{\ECExpCrit}{24}

\newcommand{\ECExpPair}{18}
\newcommand{\ECExpPairC}{20}

\newcommand{\ECExpPairTen}{9}

\newcommand{\ECExpReplaced}{30}
\newcommand{\ECExpReplacedC}{30}
\newcommand{\ECExpReplacedTen}{46}

\newcommand{\ECNeutralPair}{25}
\newcommand{\ECNeutralPairC}{22}

\newcommand{\ErhoAllOneView}{0.82}

\newcommand{\ExCapLowMean}{9}

\newcommand{\ExCapModMean}{27}

\newcommand{\ExCapSevMean}{25}

\newcommand{\ExDepressed}{30}

\newcommand{\ExOneAuc}{0.81}

\newcommand{\ExOneFlagged}{24}

\newcommand{\ExOneMissed}{14}

\newcommand{\ExOneModel}{Qwen2.5 7B}

\newcommand{\ExPersonPhq}{5}

\newcommand{\ExPersonPi}{50}

\newcommand{\ExTenLowShare}{48}

\newcommand{\ExTwoAuc}{0.80}

\newcommand{\ExTwoFlagged}{80}

\newcommand{\ExTwoMissed}{1}

\newcommand{\ExTwoModel}{Mistral 7B}

\newcommand{\LFGainDigitsC}{$-$0.005}

\newcommand{\LFGainExpC}{+0.035}

\newcommand{\LFGainFixedC}{$-$0.004}

\newcommand{\LFWinsDigitsC}{6}

\newcommand{\LFWinsExpC}{8}

\newcommand{\LFWinsFixedC}{5}
\newcommand{\LabelFreeAbove}{11}
\newcommand{\LabelFreeAboveC}{9}

\newcommand{\LabelFreeN}{11}
\newcommand{\LabelFreeNC}{11}

\newcommand{\LetterMax}{85}

\newcommand{\LockedRtwo}{0.90}

\newcommand{\ModelShare}{30.0}
\newcommand{\ModelShareC}{26.2}
\newcommand{\MxFShare}{15.8}

\newcommand{\NConfig}{880}
\newcommand{\NConfigAcc}{540}

\newcommand{\NConfigC}{440}
\newcommand{\NDefens}{44}

\newcommand{\NPart}{189}
\newcommand{\NPartC}{86}
\newcommand{\NPredEvaluated}{7}
\newcommand{\NPredEvaluatedC}{6}
\newcommand{\NPredSupported}{5}
\newcommand{\NPredSupportedC}{5}

\newcommand{\NStrict}{75}

\newcommand{\OneBestAcc}{79}

\newcommand{\OneBestModel}{Qwen2.5 14B}
\newcommand{\OneBestPrev}{43}
\newcommand{\OneBestSens}{83}

\newcommand{\PTrue}{0.18}

\newcommand{\PairAcc}{40}
\newcommand{\PairAccC}{40}

\newcommand{\PairAll}{44}

\newcommand{\PairDefens}{31}

\newcommand{\PairStrict}{12}

\newcommand{\PersonAll}{31}

\newcommand{\PersonShare}{10.5}
\newcommand{\PersonShareC}{13.0}

\newcommand{\PhiAllOneView}{0.53}

\newcommand{\PhiAllTwoView}{0.55}

\newcommand{\PhiOneOneView}{0.11}

\newcommand{\PrecFlipMax}{35}
\newcommand{\PrecFlipMed}{21}

\newcommand{\Prev}{30.2}
\newcommand{\PrevC}{33.7}
\newcommand{\PrevMax}{100}
\newcommand{\PrevMaxC}{100}

\newcommand{\PrevMin}{0}
\newcommand{\PrevMinC}{0}
\newcommand{\PrevModelMax}{99}

\newcommand{\PrevModelMin}{14}

\newcommand{\PxMShare}{8.1}

\newcommand{\RCCohortPrevPair}{19}

\newcommand{\RCLinFits}{107{,}540}
\newcommand{\RCLinFitsC}{62{,}743}
\newcommand{\RCLinNonInc}{14}

\newcommand{\RCLinPair}{16}

\newcommand{\RCLinSe}{42}

\newcommand{\RCN}{538}

\newcommand{\RCNeutralPrevPair}{24}

\newcommand{\RCNoneAcc}{61}
\newcommand{\RCNoneAccC}{59}

\newcommand{\RCNonePair}{40}
\newcommand{\RCNonePairC}{40}

\newcommand{\RCPrevAcc}{75}
\newcommand{\RCPrevAccC}{75}

\newcommand{\RCPrevEightyPair}{20}

\newcommand{\RCPrevErr}{7}
\newcommand{\RCPrevErrC}{8}
\newcommand{\RCPrevErrHi}{19}

\newcommand{\RCPrevErrLo}{3}

\newcommand{\RCPrevPair}{20}
\newcommand{\RCPrevPairC}{21}

\newcommand{\RCPrevTarget}{88}

\newcommand{\RCPrevTwentyPair}{22}

\newcommand{\RCSensPair}{27}

\newcommand{\RCSensSe}{78}

\newcommand{\RMAccExcess}{2.3}
\newcommand{\RMAccExcessC}{2.1}
\newcommand{\RMAccForced}{38}

\newcommand{\RMAccTot}{40}
\newcommand{\RMAccTotC}{40}

\newcommand{\RaterOnly}{69}

\newcommand{\RemAllNeg}{104}

\newcommand{\RemAllPos}{45}

\newcommand{\RemCov}{71}

\newcommand{\RemDeferAcc}{86}
\newcommand{\RemDeferAccC}{89}
\newcommand{\RemDeferSe}{56}
\newcommand{\RemDeferSeC}{70}

\newcommand{\RemKeptNeg}{82}

\newcommand{\RemKeptPos}{24}

\newcommand{\RemMajAcc}{78}
\newcommand{\RemMajAccC}{80}
\newcommand{\RemMajSe}{58}
\newcommand{\RemMajSeC}{65}

\newcommand{\RemMarginAcc}{82}
\newcommand{\RemMarginAccC}{82}

\newcommand{\RemSameAcc}{83}
\newcommand{\RemSameAccC}{84}

\newcommand{\RemSingleAcc}{75}
\newcommand{\RemSingleAccC}{75}

\newcommand{\RhoAlphaAuc}{0.02}
\newcommand{\RhoAlphaAucC}{$-$0.06}

\newcommand{\RhoPhqPtsd}{0.84}

\newcommand{\RhoSampAuc}{0.76}

\newcommand{\RhoXtempAuc}{0.71}
\newcommand{\RhoXtempAucC}{0.67}
\newcommand{\RtwoUniverse}{0.53}

\newcommand{\RuleEvidence}{27}

\newcommand{\RuleEvidenceN}{10}

\newcommand{\RuleGap}{28}

\newcommand{\RuleItemSumMatched}{8}

\newcommand{\RuleLetters}{21}

\newcommand{\RuleLowShare}{35}

\newcommand{\RulePtsd}{0.36}
\newcommand{\RulePtsdC}{0.23}

\newcommand{\RulePtsdPC}{0.049}

\newcommand{\RuleRtwo}{0.91}

\newcommand{\RuleSizeRho}{0.82}

\newcommand{\RuleSlope}{13}

\newcommand{\RuleVerbosity}{0.25}
\newcommand{\RuleVerbosityC}{0.26}

\newcommand{\RulesNC}{9}

\newcommand{\RulesPassedC}{9}
\newcommand{\SCAggCap}{15}

\newcommand{\SCAggRaw}{24}

\newcommand{\SCEvidenceCap}{14}

\newcommand{\SCEvidenceRaw}{21}

\newcommand{\SCLetterCap}{21}

\newcommand{\SCLetterRaw}{20}

\newcommand{\SCModelCap}{25}

\newcommand{\SCModelRaw}{35}

\newcommand{\SCPersonaCap}{8}

\newcommand{\SCPersonaRaw}{7}

\newcommand{\SCViewCap}{17}

\newcommand{\SensQone}{47}

\newcommand{\SensQthree}{100}

\newcommand{\ShareSig}{14}

\newcommand{\ShareSpecific}{4}
\newcommand{\ShareSpecificBoot}{4.3}
\newcommand{\ShareSpecificBootC}{0.9}

\newcommand{\ShareSpecificSteiger}{3.9}
\newcommand{\ShareSpecificSteigerC}{1.1}

\newcommand{\SpecQone}{5}

\newcommand{\SpecQthree}{84}

\newcommand{\TotDeltaMax}{0.008}

\newcommand{\TotModels}{four}
\newcommand{\TotParts}{25}

\newcommand{\TypeS}{16}

\newcommand{\UpLlamaLarge}{73}

\newcommand{\UpQwenLarge}{25}

\newcommand{\UpQwenSmall}{21}

\newcommand{\VarRatio}{10.6}

\newcommand{\VrDigits}{2.4}

\newcommand{\VrLodoMax}{6.3}

\newcommand{\VrLodoMin}{1.9}

\newcommand{\VrNeutral}{3.0}

\newcommand{\VrReg}{2.8}

\newcommand{\WidthLab}{25}

\newcommand{\WidthPpiFin}{27}

\newcommand{\XtempWithin}{0.67}
\newcommand{\XtempWithinC}{0.59}

\graphicspath{{figures/}}

\title{Language-model ratings of depression reflect\\
the rater more than the patient}

\author{Baihan Lin$^{1,2,3,4,5}$\\[3pt]
  \parbox{0.92\textwidth}{\centering\normalfont\small
  $^1$Department of Artificial Intelligence and Human Health, Icahn School of Medicine at Mount Sinai, New York, NY, USA\\
  $^2$Department of Psychiatry, Icahn School of Medicine at Mount Sinai, New York, NY, USA\\
  $^3$Department of Neuroscience, Icahn School of Medicine at Mount Sinai, New York, NY, USA\\
  $^4$Mental Illness Research, Education and Clinical Center, James J. Peters VA Medical Center, Bronx, NY, USA\\
  $^5$Berkman Klein Center for Internet \& Society, Harvard University, Cambridge, MA, USA\\[3pt]
  Correspondence: \texttt{baihan.lin@mssm.edu}}}

\date{}

\renewcommand{\shorttitle}{Language-model ratings reflect the rater}
\renewcommand{\headeright}{}
\renewcommand{\undertitle}{Preprint}
\hypersetup{pdftitle={Language-model ratings of depression reflect the rater more than the patient}, pdfauthor={Baihan Lin}}

\begin{document}
\maketitle

\begin{abstract}
Depression has no diagnostic blood test. Language models promise tireless, consistent assessment, but can accurate raters disagree about individuals? We pre-registered \NConfig{} language-model raters, crossing 11 open models with prompting and scoring choices, and applied them to \NPart{} interviews against the eight-item Patient Health Questionnaire. Model choice explained \ModelShare\% of summed-symptom score variance, stable participant differences \PersonShare\%. Two randomly drawn raters with area under the receiver operating characteristic curve (AUC) $\geq$ 0.70 disagreed on screening decisions for \PairAcc\% of participants, on average. Average over-rating governed how many were flagged, yet equal-capacity raters chose differently for about one participant in five. A locked analysis of \NPartC{} new interviews reproduced the main pre-registered findings. Exploratory recalibration with 40 labelled participants raised accuracy from about 60\% to 75\% and halved disagreement, leaving one participant in five decided differently. Calibration repaired much of the rater dependence without securing agreement about individuals.
\end{abstract}

\section*{Introduction}

A depression rating can shape who is offered further assessment. If the interview stays the same, how much should the decision depend on the software reading it? Psychiatry has confronted a version of this question before. A 1971 study found that American and British psychiatrists disagreed after watching the same videotaped interviews\cite{Kendell1971}. Agreement was subsequently found to be poor for most diagnostic categories\cite{Spitzer1974}. Psychiatry responded with explicit criteria and structured interviews\cite{Feighner1972,Spitzer1978}. Half a century later, large language models (LLMs) promise scalable, inexpensive, repeatable ratings from clinical conversations\cite{Obradovich2024,Stade2024,Demszky2023,Galatzer2023}. Early evaluations are encouraging: open models achieved clinically acceptable total-score error on structured psychiatric interviews\cite{Kebe2026}, reached intraclass correlations near 0.75 with trained raters of psychosis-risk and schizophrenia-spectrum interviews\cite{Zhu2026psychosis,Chen2026bprs}, and recovered self-reported depression severity from transcripts\cite{Rosenman2024,Teferra2025}. Psychotherapy dialogue also supports alliance inference\cite{Lin2023alliance,Lin2025compass} and topical analysis\cite{Lin2023topics}.

But a model becomes a rater only through choices. Someone must choose the model and word the request, perhaps as a questionnaire, in a psychiatrist's voice or as the patient answering about themselves. They must choose digits or letters and their order, decide whether the model reads the patient's words or the whole conversation, and turn its answer into a score. These choices can define plausible raters, although evaluations rarely show the alternatives. Outputs shift with prompt formatting\cite{Sclar2024,Mizrahi2024}, answer order and labels\cite{Zheng2024,Tjuatja2024}, inference settings\cite{Pugh2024}, and even supposedly deterministic settings\cite{Atil2024}. Such choices reverse social-science conclusions\cite{Baumann2025}, alter depression inference\cite{Lorenzoni2025}, change diagnoses when clinical cases are reworded\cite{Alkaeed2026}, and shift severity judgements across languages\cite{XuHu2026}. High self-consistency makes reliability easier to establish than validity\cite{Norman2026,Batzdorfer2026,LinZ2026}. For a clinic, the uncertainty concerns whose assessment changes when one plausible rater replaces another, and how much local labels can repair.

Generalizability theory asks whether differences between people survive changes in how they are measured. It separates contributions from the person and the measurement facets\cite{Cronbach1972,Brennan2001}, with dependable scores requiring the person's contribution to dominate error. Defensible configurations define a multiverse\cite{Steegen2016,Simonsohn2020}. Disagreement among similarly accurate configurations is predictive multiplicity\cite{Marx2020,Black2022}. Even predictors that agree on average can induce different decisions, motivating recalibration for downstream decisions\cite{Du2025}. Calibration, long a weak point of clinical prediction\cite{VanCalster2019}, connects measurement to action. Together, these ideas distinguish whether a rater orders people well, whether its threshold flags an appropriate number, and whether another defensible rater selects the same people.

We crossed 11 open models, five wordings, two conversation views, two answer-reading rules and four score constructions (eight symptom ratings in three formats, or one overall rating), creating \NConfig{} raters for \NPart{} interviews with the eight-item Patient Health Questionnaire (PHQ-8) as criterion\cite{Kroenke2009}. Analysis code was fixed at pre-registration before comparison with the criterion\cite{Nosek2018,MacCoun2015}. The locked procedure was then applied once to \NPartC{} new interviews from a related corpus, with an autonomous interviewer and automatic transcription. The rater outweighed the patient in score variance, and accurate rankings concealed divergent decisions. Offsets governed how many were flagged, but equal-capacity raters still chose different people. Even answer letters mattered. We then examined how far local recalibration could repair these differences: decisions improved, while disagreement about individuals remained.

\section*{Results}

\subsection*{The same interview, opposite verdicts}

\begin{figure}[t]
  \centering
  \includegraphics[width=\textwidth]{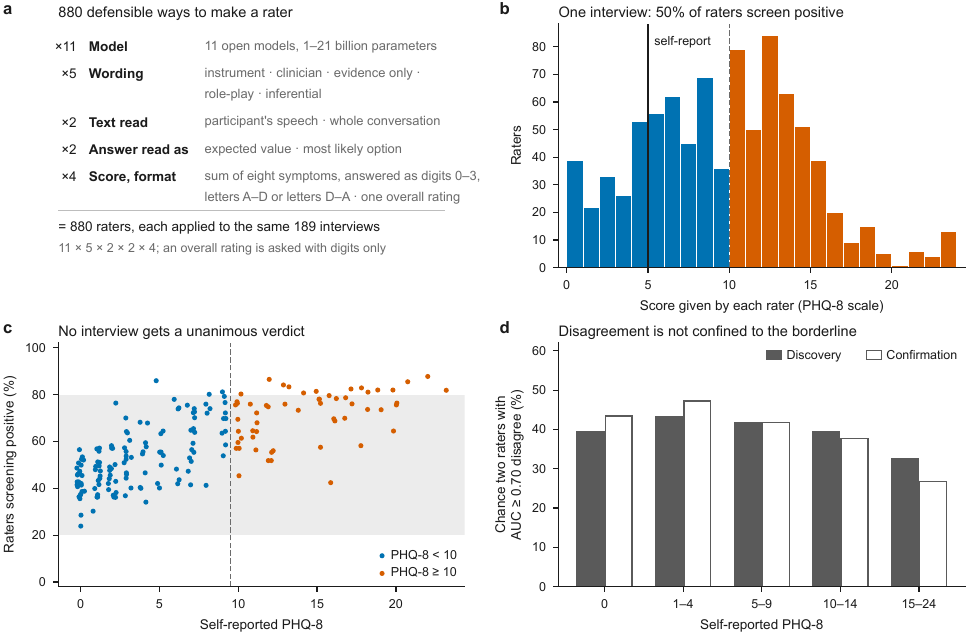}
  \caption{\textbf{The same interview, opposite verdicts.} \textbf{a}, The choices that make a rater; each of the \NConfig{} combinations was applied to every interview. \textbf{b}, Scores given to one participant by all \NConfig{} raters, on the PHQ-8 scale; black line, the participant's self-reported PHQ-8; dashed line, the screening threshold of 10; orange, scores of 10 or more. \textbf{c}, For every participant, the share of raters screening them positive, against their self-reported PHQ-8 (jittered horizontally); grey band, 20--80\% of raters; dashed line, between self-reported totals of 9 and 10. \textbf{d}, The chance that two different raters with AUC $\geq$ 0.70, drawn at random, give a participant opposite decisions, by self-reported severity; dark grey bars, discovery; white bars, confirmation.}
  \label{fig:verdicts}
\end{figure}

One participant reported mild symptoms (PHQ-8 \ExPersonPhq{}), below the screening threshold of 10, yet \ExPersonPi\% of \NConfig{} raters screened them positive (Fig.~\ref{fig:verdicts}b). No participant received a unanimous verdict, and \ContAll\% split raters between 20\% and 80\% (Fig.~\ref{fig:verdicts}c). Some dissent becomes likelier as raters accumulate: from \DiscKTwo\% of participants with two AUC $\geq$ 0.70 raters to \DiscKTen\% with ten (Extended Data Fig.~\ref{fig:ed-grid}a).

We therefore measured the chance that two distinct, randomly drawn raters disagree: \PairAll\% across all raters, \PairAcc\% among the \NConfigAcc{} with AUC $\geq$ 0.70 (95\% participant interval \BootPairLo--\BootPairHi\%, raters fixed), and \PairStrict\% among the \NStrict{} with AUC, sensitivity and specificity all at least 0.70. Even this strict set gave opposite decisions to \DiscStrict\% of participants (Extended Data Fig.~\ref{fig:ed-grid}b). An outcome-independent, clinically defensible core using instrument or psychiatrist wording, digits, expected values and symptom sums (\NDefens{} raters, AUC \DefensAucMin{}--\DefensAucMax{}) disagreed on \PairDefens\% (Supplementary Information). Nor were only borderline cases affected (Fig.~\ref{fig:verdicts}d). Participants reporting few or no symptoms (PHQ-8 $\leq$ 4) were flagged by \RuleLowShare\% of AUC $\geq$ 0.70 raters, which disagreed about them as much as about anyone.

\subsection*{Accuracy does not decide who is flagged}

\begin{figure}[t]
  \centering
  \includegraphics[width=\textwidth]{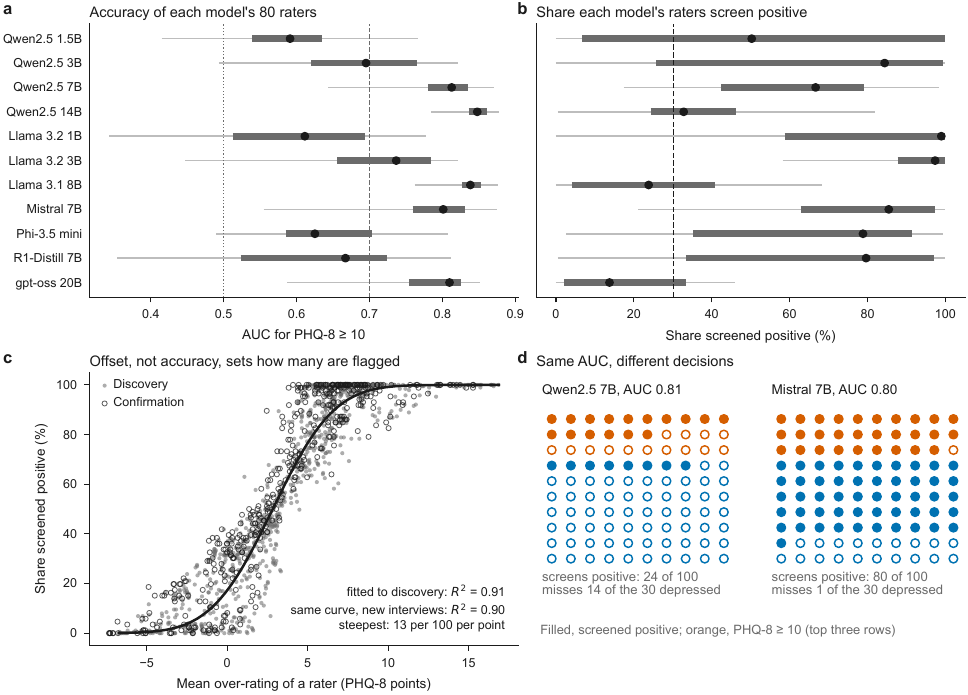}
  \caption{\textbf{Accuracy does not decide who is flagged.} \textbf{a}, AUC of each model's 80 raters: dot, median; thick line, interquartile range; thin line, range. Dotted line, chance; dashed line, 0.70. \textbf{b}, Share of participants that each model's raters screen positive; dashed line, self-reported prevalence. \textbf{c}, Share screened positive against the rater's mean over-rating of the PHQ-8, for every rater in discovery (filled) and confirmation (open); line, normal-ogive curve fitted to the discovery raters, with its $R^2$ in discovery and, applied unchanged, in confirmation. \textbf{d}, Decisions for 100 people with the sample's prevalence by the typical configuration (median AUC) of two models with nearly the same AUC; shown is the pair of models whose typical configurations differ in AUC by at most 0.02 and differ most in the share screened positive.}
  \label{fig:decisions}
\end{figure}

Many raters met published standards of accuracy. AUC asks how well scores rank participants with a positive screen above those without one, without choosing a threshold. AUC for PHQ-8 $\geq$ 10 ranged from \AucMin{} to \AucMax{} (median \AucMed{}), with \AucBelowSeventy\% below 0.70 (Fig.~\ref{fig:decisions}a; Extended Data Fig.~\ref{fig:ed-multiverse}). Between-rater AUC variance was \VarRatio{} times its sampling variance: rater choice mattered more than participant sampling. Decisions told another story. Raters flagged \PrevMin--\PrevMax\% of participants against \Prev\% self-reported prevalence; model medians ranged from \PrevModelMin\% to \PrevModelMax\% (Fig.~\ref{fig:decisions}b).

The rater's offset, its mean over-rating of PHQ-8, governed the share flagged along a normal-ogive curve ($R^2$ = \RuleRtwo{}; Fig.~\ref{fig:decisions}c). Applied unchanged to the new interviews, that curve explained \LockedRtwo{} of the variance. The relation is partly built in: offset and threshold crossings summarise the same scores. Nevertheless, location nearly exhausted what mattered at the threshold, and AUC cannot see location. Near the curve's centre, each point of over-rating flagged \RuleSlope{} additional people per 100. A rater behaves like a bathroom scale with a misplaced zero: it may rank people well, but its offset decides who crosses the line. \ExOneModel{} and \ExTwoModel{}, in their typical configurations, had almost identical AUCs (\ExOneAuc{}, \ExTwoAuc{}). Per 100 people like those studied, the first would flag \ExOneFlagged{} and miss \ExOneMissed{} of \ExDepressed{} positive screens; the second would flag \ExTwoFlagged{} and miss \ExTwoMissed{} (Fig.~\ref{fig:decisions}d). AUC $\geq$ 0.70 raters differing by at most 0.02 in AUC differed by a median \RuleGap{} percentage points in the share flagged. Across all raters, sensitivity quartiles were \SensQone--\SensQthree\%, specificity \SpecQone--\SpecQthree\%.

\subsection*{The rater outweighs the patient}

\begin{figure}[t]
  \centering
  \includegraphics[width=\textwidth]{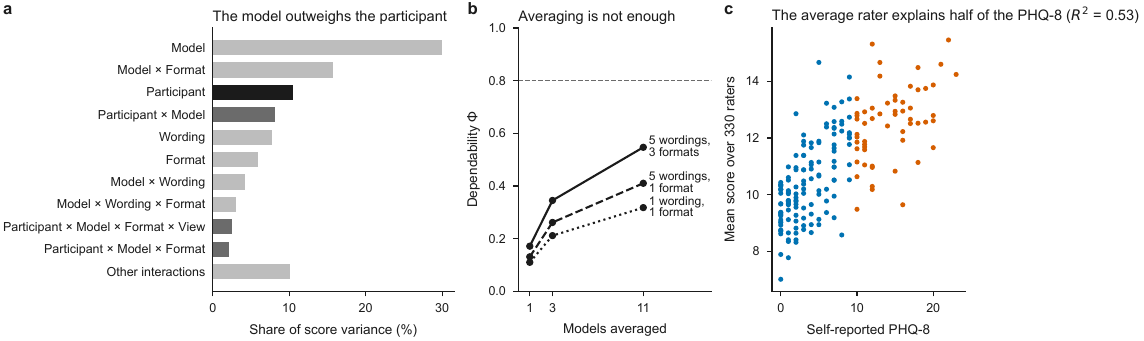}
  \caption{\textbf{The rater outweighs the patient.} \textbf{a}, Shares of the variance of the expected item sum in the generalizability analysis of the 330 expected-value item-sum raters (discovery); black, participant main effect; dark grey, interactions involving the participant; light grey, components of the rater alone. \textbf{b}, Dependability of a score averaged over models, for one or all wordings and formats, with both views of the conversation averaged; dashed line, 0.80. \textbf{c}, Participants' mean score over the same 330 raters against their self-reported PHQ-8; orange, PHQ-8 of 10 or more.}
  \label{fig:rater}
\end{figure}

Each interview was held fixed while the rater changed. Stable participant differences, the person main effect, explained \PersonShare\% of summed-symptom score variance. Model choice explained \ModelShare\%, or \VrReg{} times as much (Fig.~\ref{fig:rater}a). Model-by-format interaction explained \MxFShare\%: formats shifted models differently. Participant-by-model interaction explained \PxMShare\%: models scored participants differently relative to one another, reflecting rankings or score scales. Crediting every participant-related component to the participant still gave \PersonAll\%, against \RaterOnly\% for the rater alone. The ordering survived restriction to digits (model-to-person ratio \VrDigits{}), one neutral wording, format and view (\VrNeutral{}), and omission of each developer's models (\VrLodoMin{}--\VrLodoMax{}). Dependability was \PhiOneOneView{} for one model, wording, format and view. Averaging 11 models, five wordings and three formats raised it to only \PhiAllOneView{}, or \PhiAllTwoView{} with both views (Fig.~\ref{fig:rater}b; Extended Data Fig.~\ref{fig:ed-dstudy}). Participants' mean scores across the 330 expected-value item-sum raters explained \RtwoUniverse{} of PHQ-8 variance (Fig.~\ref{fig:rater}c). Even the average rater captured only part of the questionnaire.

\subsection*{Small choices move the threshold}

\begin{figure}[t]
  \centering
  \includegraphics[width=\textwidth]{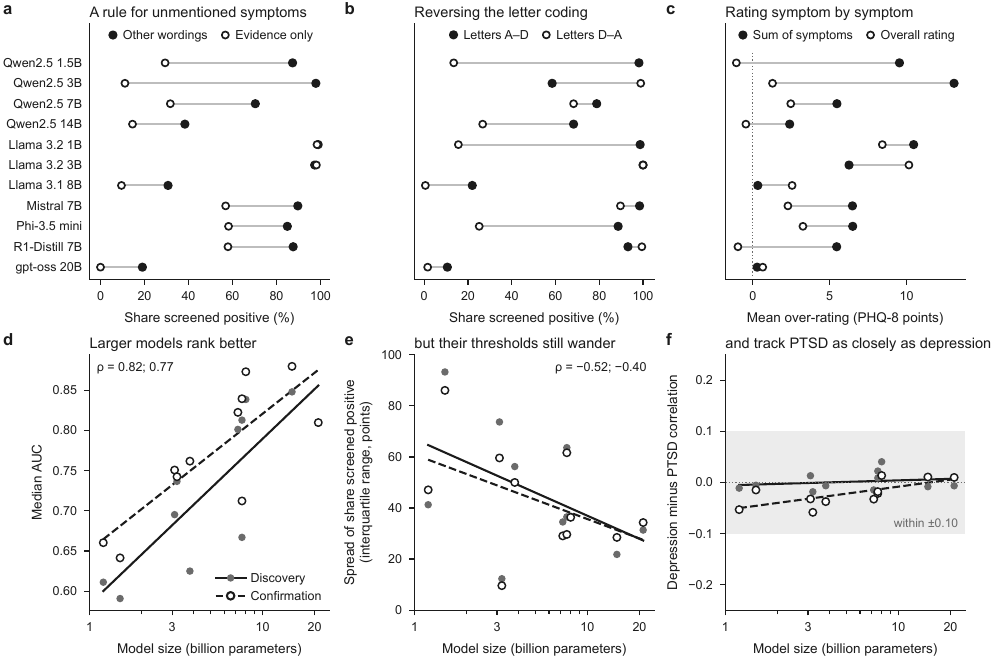}
  \caption{\textbf{Small choices move the threshold, and size buys ranking but not calibration or specificity.} \textbf{a}--\textbf{c}, For each model, the median share screened positive with the other wordings (filled) and with the instruction to rate unmentioned symptoms as absent (open) (\textbf{a}), and with answer letters A--D meaning `not at all' to `nearly every day' (filled) and the reverse (open) (\textbf{b}); and the median over-rating with digit answers when symptoms are summed (filled) or the total is rated directly (open) (\textbf{c}). \textbf{d}--\textbf{f}, Against model size, one point per model: median AUC (\textbf{d}), interquartile range across a model's raters of the share screened positive (\textbf{e}), and the median difference between a rater's correlation with the PHQ-8 and with PTSD severity (\textbf{f}); grey band, differences within $\pm$0.10. Lines, least-squares fits against log size; filled markers and solid lines, discovery; open markers and dashed lines, confirmation; $\rho$, Spearman correlations across models.}
  \label{fig:dials}
\end{figure}

Apparent technicalities moved the threshold (Fig.~\ref{fig:dials}). Rating only explicitly reported symptoms, an assessment-policy change from the other four instructions, lowered the median share flagged by \RuleEvidence{} percentage points (\RuleEvidenceN{} of 11 models; Fig.~\ref{fig:dials}a). Reversing answer letters so A meant `nearly every day' instead of `not at all', with no clinical change, moved it by a median \RuleLetters{} points and up to \LetterMax{} (Fig.~\ref{fig:dials}b). The interview and the symptom descriptions were unchanged. With digits, symptom sums over-rated relative to direct totals in \RuleItemSumMatched{} of 11 models in both sets (Fig.~\ref{fig:dials}c), despite similar discrimination (median AUC \AucItemSumMed{}, \AucDirectMed{}). For symptom sums, digits outperformed letters (\AucDigitsMed{} versus \AucLettersMed{}, or \AucLettersRevMed{} reversed). Larger models ranked better (Spearman correlation \RuleSizeRho{}; Fig.~\ref{fig:dials}d), without better calibration or specificity. Their configurations' decisions converged, but even the largest models varied in share flagged by tens of percentage points (Fig.~\ref{fig:dials}e).

Deployment choices also changed decisions. Running Qwen2.5 7B as a 4-bit file with another inference library changed a median \PrecFlipMed\% of decisions (maximum \PrecFlipMax\%) under identical configurations: a comparison of deployment stacks, not compression alone. Automatic instead of manual transcripts changed \AsrFlipMed\% (maximum \AsrFlipMax\%). Increasing model size with other choices fixed changed \UpQwenSmall--\UpQwenLarge\% within Qwen2.5 and \UpLlamaLarge\% from Llama 3.2 3B to Llama 3.1 8B.

\subsection*{What the raters read}

\begin{figure}[t]
  \centering
  \includegraphics[width=\textwidth]{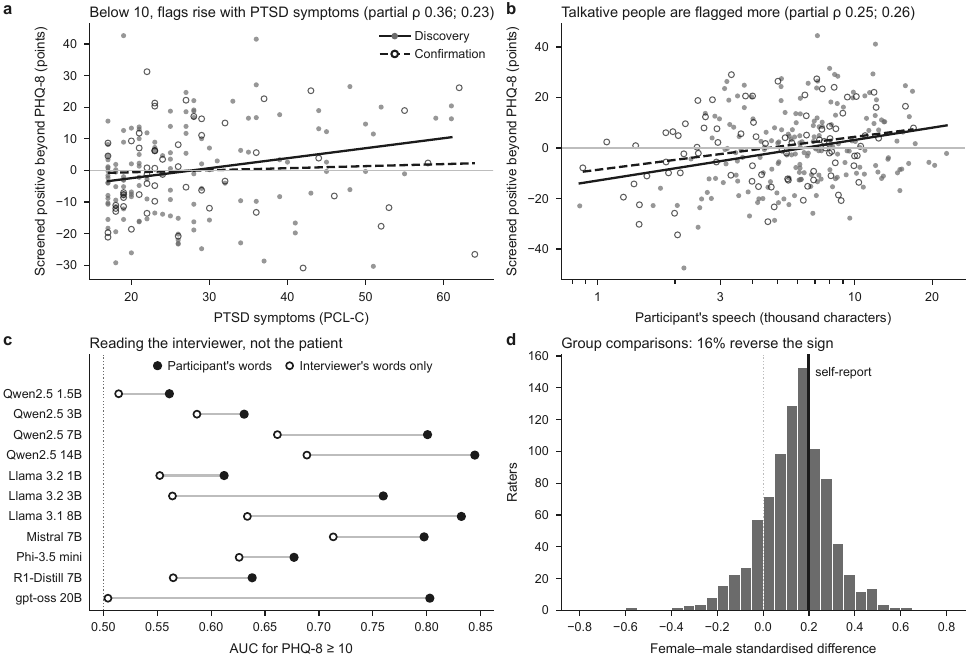}
  \caption{\textbf{What the raters read.} \textbf{a}, Among participants below the screening threshold, the share of raters with AUC $\geq$ 0.70 screening them positive beyond what their PHQ-8 predicts (residual of a linear fit, percentage points), against PTSD symptoms (PCL-C). \textbf{b}, The same for all participants, against the amount of speech (logarithmic scale). Points and lines show residuals and linear fits on the original scales; the statistics in the panel titles are partial Spearman correlations on ranks (discovery; confirmation). Filled markers and solid lines, discovery; open markers and dashed lines, confirmation. \textbf{c}, AUC of each model from the participant's speech (filled) and from the interviewer's speech alone (open), median over wordings and formats of the expected-value item sum (discovery). \textbf{d}, Female--male standardised difference estimated by every rater (discovery); black line, the difference in self-reported PHQ-8.}
  \label{fig:who}
\end{figure}

At every model size, scores showed little preferential association with PHQ-8 over PTSD severity (Fig.~\ref{fig:dials}f). Only \ShareSpecific\% of raters correlated significantly more strongly with PHQ-8 (\ShareSpecificBoot\% by paired bootstrap). The criteria themselves correlated strongly ($\rho$ = \RhoPhqPtsd{}), preventing conclusions about diagnostic separation. Below the screening threshold, higher PTSD symptoms predicted more flags than PHQ-8 alone predicted (partial Spearman \RulePtsd{}; confirmation \RulePtsdC{}, $P$ = \RulePtsdPC{}; Fig.~\ref{fig:who}a). Talkativeness showed the same association (\RuleVerbosity{}, \RuleVerbosityC{}; Fig.~\ref{fig:who}b). These associations warrant specificity tests without identifying what drives ratings. Interviewer speech alone, containing no participant report, yielded median model AUC \AucInterviewer{} (Fig.~\ref{fig:who}c): above chance, as follow-up questions predict\cite{Burdisso2024}, but below the predicted 0.60. Group conclusions were also fragile. The self-reported female--male difference was small (standardised difference \DTrue{}, $P$ = \PTrue{}), yet \TypeS\% of raters reversed its sign and \ShareSig\% reported a nominally significant difference absent from self-reports (Fig.~\ref{fig:who}d).

\subsection*{Checks that need no labels}

\begin{figure}[t]
  \centering
  \includegraphics[width=\textwidth]{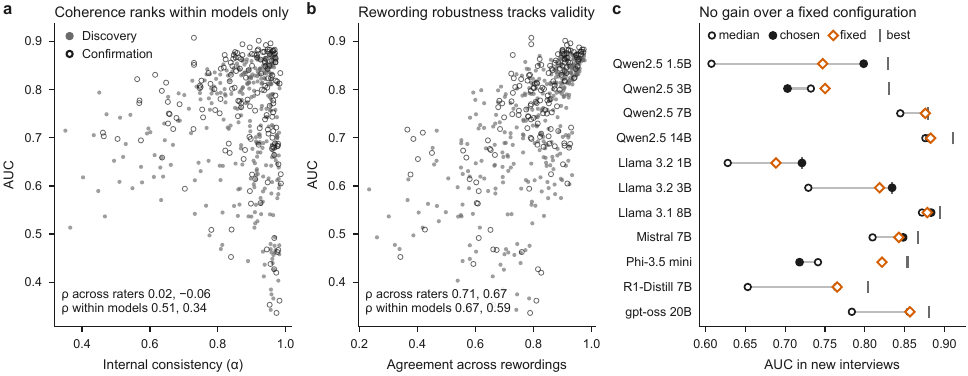}
  \caption{\textbf{Checks that need no labels.} \textbf{a},\textbf{b}, AUC of each item-sum rater against its internal consistency (\textbf{a}) and against its agreement with the same model and format under the other wordings (\textbf{b}); filled, discovery; open, confirmation; Spearman correlations across all raters and medians of the correlations within models, for discovery and then confirmation. \textbf{c}, In the confirmation set, the AUC of each model's median item-sum configuration, of the configuration chosen without labels by agreement across rewordings, of a fixed neutral configuration (the questionnaire's wording, digit answers, the participant's speech, expected value) and of its best configuration; the rule was specified before the confirmation analysis.}
  \label{fig:reassurance}
\end{figure}

These raters were consistent, as evaluations commonly report (median Cronbach's $\alpha$ \AlphaMed{}). Internal consistency was unrelated to validity across raters (Spearman correlation with AUC \RhoAlphaAuc{}; confirmation \RhoAlphaAucC{}; Fig.~\ref{fig:reassurance}a), although informative within models (median correlation \AlphaWithin{}, \AlphaWithinC{}). A model can distribute one global impression coherently across symptoms whether that impression is right or wrong. Agreement across rewordings tracked validity both across raters and within models (\RhoXtempAuc{}, \XtempWithin{}; confirmation \RhoXtempAucC{}, \XtempWithinC{}; Fig.~\ref{fig:reassurance}b). Agreement between independent runs sampled at temperature 1 also tracked validity (\RhoSampAuc{} across raters).

Before confirmation, we specified a label-free rule choosing each model's configuration with greatest agreement across rewordings. As predicted, it beat the model median in \LabelFreeAboveC{} of \LabelFreeNC{} confirmation models (\LabelFreeAbove{} of \LabelFreeN{} in discovery). But the rule chose only expected-value configurations, whereas its comparator included cruder most-likely-option scores. Against eligible configurations' median, it won in \LFWinsExpC{} of 11 models (mean AUC gain \LFGainExpC{}). Against the digit-answer median it won in \LFWinsDigitsC{} (\LFGainDigitsC{}), and against a fixed neutral configuration (instrument wording, digits, participant speech, expected values) in \LFWinsFixedC{} (\LFGainFixedC{}; Fig.~\ref{fig:reassurance}c). Avoiding letters explained most of the gain. Rewording robustness can flag fragile, often invalid configurations, but found none better than a sensible default.

\subsection*{How many versus who}

\begin{figure}[t]
  \centering
  \includegraphics[width=\textwidth]{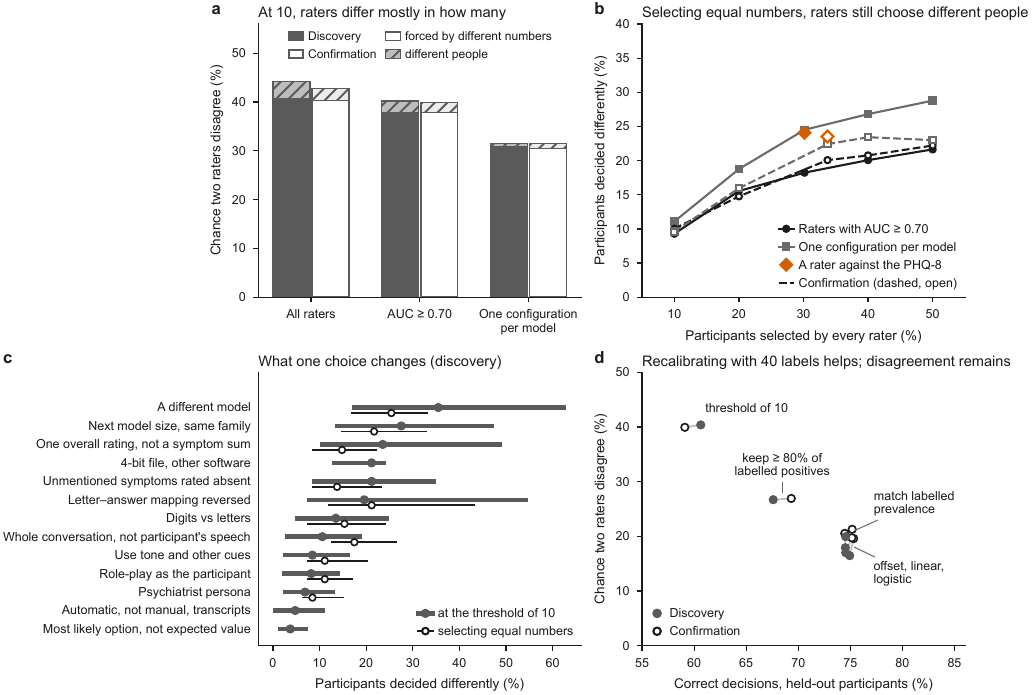}
  \caption{\textbf{How many versus who.} \textbf{a}, The chance that two different raters disagree about a participant at the threshold of 10, split by the identity $P(A \neq B) = |p_A - p_B| + 2\min\{P(A{=}1, B{=}0), P(A{=}0, B{=}1)\}$ into the part forced by their different positive rates (solid) and the rest, from flagging different people (hatched), for all raters, raters with AUC $\geq$ 0.70 and one fixed configuration per model; dark grey, discovery; white, confirmation. \textbf{b}, Participants decided differently when every rater selects the same share of participants, its highest-scoring ones, by two expected-value raters with AUC $\geq$ 0.70 or two fixed configurations (one per model); diamonds, participants decided differently by a rater and the self-reported criterion when the rater selects as many as screen positive. Filled markers and solid lines, discovery; open markers and dashed lines, confirmation. \textbf{c}, Participants decided differently by configurations that differ in one choice only (discovery; changes of wording are relative to the questionnaire's own wording): dot, median over pairs of configurations; bar, interquartile range; filled, at the threshold of 10; open, selecting as many as screen positive (expected-value scores only); deployment-stack and transcript comparisons are for decisions only. \textbf{d}, For raters admitted by their AUC on 40 labelled participants and recalibrated with the same 40 labels, the chance that two raters disagree against the share of correct decisions, both on the held-out participants (means over 200 splits); filled, discovery; open, confirmation. Exploratory analyses.}
  \label{fig:whowhat}
\end{figure}

Disagreement can concern how many people to flag or which people. An identity separates the minimum disagreement forced by unequal positive rates from the remainder: total disagreement equals the absolute rate difference plus twice the smaller one-sided disagreement. At the screening threshold of 10, unequal rates dominated. Among AUC $\geq$ 0.70 raters, \RMAccTot\% disagreement comprised \RMAccForced{} percentage points forced by rates and \RMAccExcess{} beyond them (confirmation \RMAccTotC\% total, \RMAccExcessC{} excess; Fig.~\ref{fig:whowhat}a). All raters, the defensible core and one fixed configuration per model showed the same pattern. Decisions were nearly nested: lenient raters flagged almost everyone strict raters flagged, then added others.

Nested decisions need not imply shared severity rankings. To separate who is selected from how many, we gave each rater the same number of places to fill. Selecting equal numbers, matching self-reported prevalence, expected-value raters with AUC $\geq$ 0.70 still chose differently for \ECExpPair\% of participants (95\% interval \BootCapLo--\BootCapHi\%), replacing \ECExpReplaced\% of each other's selections (confirmation \ECExpPairC\%, \ECExpReplacedC\%; Fig.~\ref{fig:whowhat}b). Disagreement was \ECDefensPair\% within the defensible core (\ECDefensPairC\% confirmation) and \ECNeutralPair\% with one fixed configuration per model (\ECNeutralPairC\%), without accuracy-based selection. Selecting the top tenth reduced disagreement to \ECExpPairTen\% because fewer people were selected, while still replacing \ECExpReplacedTen\% of selections. Raters disagreed with the criterion on \ECExpCrit\%, more than with one another, indicating shared departures from PHQ-8. The two forms of disagreement also affected different people. At the threshold of 10, participants reporting few or no symptoms contributed almost half (\ExTenLowShare\%). At equal capacity, their exchange probability was only \ExCapLowMean\%, compared with \ExCapModMean\% at PHQ-8 10--14 and \ExCapSevMean\% at 15 or above (Supplementary Information).

Changing one choice isolates its consequences (Fig.~\ref{fig:whowhat}c). Changing models altered a median \SCModelRaw\% of decisions at 10 and \SCModelCap\% at equal capacity. Reversing letters altered \SCLetterRaw\% and \SCLetterCap\%: clinically meaningless relabelling changed both how many and who. The psychiatrist persona changed least among wordings (\SCPersonaRaw\%, \SCPersonaCap\%). Full dialogue rather than participant speech changed \SCViewCap\% of equal-capacity selections. The unmentioned-symptom rule and direct-total scoring chiefly moved thresholds: \SCEvidenceRaw\% versus \SCEvidenceCap\%, and \SCAggRaw\% versus \SCAggCap\%, respectively.

The dominance of offsets suggested a practical repair: use local labels to reset the thresholds. This repaired much of the disagreement. Across 200 random splits, 40 labelled participants admitted raters with AUC $\geq$ 0.70 (mean \RCN{} of \NConfig{}) and recalibrated them by five rules: subtracting mean over-rating, linear regression, logistic regression, the attainable threshold closest to labelled prevalence, or the highest threshold retaining at least 80\% of labelled positives. Evaluation used the remaining participants, whose labels informed no step. Prevalence matching raised accuracy from \RCNoneAcc\% to \RCPrevAcc\% (confirmation \RCNoneAccC\% to \RCPrevAccC\%) and halved disagreement, from \RCNonePair\% to \RCPrevPair\% (\RCNonePairC\% to \RCPrevPairC\%; Fig.~\ref{fig:whowhat}d). Every rule improved accuracy.

Lower disagreement sometimes meant more missed cases: linear recalibration gave \RCLinPair\% disagreement and median sensitivity \RCLinSe\%. Retaining 80\% of labelled positives gave held-out sensitivity \RCSensSe\% and disagreement \RCSensPair\%. Each rule sets a cut-off on the rater's score. Increasing fits preserve rankings, as did all but \RCLinNonInc{} of \RCLinFits{} linear fits. Residual disagreement after prevalence matching remained near one in five, resembling equal-capacity disagreement with all labels available. It persisted with 20 or 80 labels (\RCPrevTwentyPair\%, \RCPrevEightyPair\%), fixed configurations (\RCNeutralPrevPair\%) and whole-cohort admission (\RCCohortPrevPair\%; Extended Data Fig.~\ref{fig:ed-recal}).

Prevalence matching brought the share flagged within a median \RCPrevErr{} percentage points of self-reported prevalence (confirmation \RCPrevErrC{}), with a central 95\% split-to-split range of \RCPrevErrLo--\RCPrevErrHi{} points. Score ties sometimes prevented matching labelled prevalence within 5 points (achieved in \RCPrevTarget\% of cases). For prevalence estimation, the 40 labels alone gave intervals nearly as narrow as prediction-powered estimates using ratings\cite{Angelopoulos2023} (\WidthLab{} versus \WidthPpiFin{} percentage points), with similar coverage (Extended Data Fig.~\ref{fig:ed-remedies}c).

Calibration did not always travel. We transferred discovery calibrations unchanged for configurations with discovery AUC $\geq$ 0.70, from manual transcripts and operator-controlled interviews to automatic transcripts and autonomous interviews. Subtracting discovery offsets over-corrected: median share flagged fell from \CarryNonePrev\% to \CarryOffPrev\%, against \CarryTruePrev\% prevalence. Prevalence-matched thresholds transferred reasonably (\CarryPrevThPrev\% flagged, median sensitivity \CarryPrevThSens\%). The best discovery configuration (\OneBestModel{}), with its discovery threshold, flagged \OneBestPrev\%, with \OneBestAcc\% accuracy and \OneBestSens\% sensitivity. In this single related transfer, transportability depended on the calibration and required checking.

Could disagreement guide deferral? Pooling and abstention have been proposed for medical predictive multiplicity\cite{Lockfisch2025}, but selective prediction can fail particular groups\cite{LechugaLopez2026}. We compared rules on the same held-out participants at matched coverage (Extended Data Fig.~\ref{fig:ed-defer}). Majority accuracy was \RemMajAcc\%, versus \RemSingleAcc\% for a single recalibrated rater (confirmation \RemMajAccC\%, \RemSingleAccC\%). Requiring 80\% consensus retained \RemCov\% of participants, with majority accuracy \RemDeferAcc\% (\RemDeferAccC\%). Single raters achieved \RemSameAcc\% on those participants (\RemSameAccC\%), or \RemMarginAcc\% (\RemMarginAccC\%) when deferring their own closest calls at equal coverage. Consensus selection outperformed a rater's own margin, but most apparent gain came from excluding difficult participants, disproportionately positive screens: \RemKeptPos{} of \RemAllPos{} positives remained, versus \RemKeptNeg{} of \RemAllNeg{} negatives. Majority sensitivity changed from \RemMajSe\% overall to \RemDeferSe\% among retained participants (confirmation \RemMajSeC\% to \RemDeferSeC\%). These sensitivities describe different groups. Deferred participants are neither detected nor missed until reviewed. Without costs for review and missed cases, these comparisons do not define a triage policy.

\subsection*{The regularities held in new interviews}

The locked procedure was applied once to \NPartC{} E-DAIC interviews (sessions 600--718) from the same laboratory, with an autonomous interviewer and automatic transcription (\NConfigC{} participant-speech raters). AUC ranged from \AucMinC{} to \AucMaxC{} (median \AucMedC{}), and share flagged from \PrevMinC\% to \PrevMaxC\%, against \PrevC\% prevalence (Extended Data Fig.~\ref{fig:ed-confirm}). AUC $\geq$ 0.70 raters disagreed on \PairAccC\%. Participant differences explained \PersonShareC\% of variance, model choice \ModelShareC\%. Of \NPredEvaluatedC{} evaluable pre-registered predictions, \NPredSupportedC{} were supported. As in discovery, 40-label calibration did not halve prevalence spread. Discovery supported \NPredSupported{} of \NPredEvaluated{} predictions, also falling short on interviewer speech. All \RulesPassedC{} of \RulesNC{} discovery regularities met criteria specified before confirmation (Extended Data Table~\ref{tab:rules}).

\section*{Discussion}

A language model asked to rate depression does not report a property of the patient alone. In these interviews, model choice explained about three times more summed-symptom score variation than stable differences between participants. Raters that ranked participants well still disagreed about many of them, and how many people were flagged depended on the offset rather than the accuracy. Psychiatry spent decades responding to unreliable diagnosis with explicit criteria and structured interviews\cite{Kendell1971,Spitzer1974,Feighner1972,Spitzer1978}. Automation promises relief from variation between human raters. Here, the rater returns as a configuration: a model, an instruction and a choice of letters, often undocumented. Choices made when building an instrument can become differences in how it judges a person.

Earlier findings of accurate, consistent LLM ratings\cite{Kebe2026,Zhu2026psychosis,Chen2026bprs} remain compatible with ours: they usually examined few configurations, and our best were accurate too. Ranking versus calibration\cite{VanCalster2019}, predictive multiplicity\cite{Marx2020,Black2022} and downstream decision disagreement\cite{Du2025} are established distinctions. The multiverse makes these distinctions tangible for one clinical decision. It measures how much a decision about a person depends on choices that an evaluation usually leaves in the background, and how much of that dependence a simple local repair removes. Clinics act at thresholds, where AUC is blind to offsets. Internal consistency did not rank configurations across models, and rewording robustness identified fragility without improving on a neutral default. For prevalence estimation, ratings added little precision beyond the 40 questionnaire labels alone in this cohort. The value of automation therefore needs to be established against what the same local measurement effort can achieve without it. Alliance estimates also inform classification\cite{Lin2024wat} and conversational recommendations\cite{Lin2023supervisor,Lin2023companion,Gunal2024decision}.

Calibration exposed the distinction between agreement about populations and agreement about people. A few dozen local labels repaired much of the disagreement: decisions became more accurate and more alike, and prevalence estimates became better calibrated. But agreement about how many is not agreement about who. An increasing recalibration leaves a rater's ranking untouched, and different configurations ranked participants differently enough to make opposite selection decisions for about one participant in five even when selecting equal numbers. Resetting a scale's zero cannot reconcile two scales that disagree about who is heavier. In a service using these ratings to allocate a fixed number of assessment appointments, the implication would be different names on a referral list of the same length.

Pooling improved accuracy; consensus deferral improved it further at matched coverage, while disproportionately excluding positive screens. Decision-focused reconciliation\cite{Du2025} may narrow the residual. Until demonstrated, individual agreement must be measured and calibration checked locally. Agreement itself is also insufficient: raters shared departures from PHQ-8, and choosing a threshold trades missed positive screens against unnecessary flags. The clinical objective must therefore be specified alongside the calibration rule.

The ratings also carried information beyond depression, raising the question of what an apparently successful score measures. They tracked PTSD as closely as PHQ-8. A preprint on these corpora found symptom geometry closer to questionnaire wording than self-report\cite{Lin2026compass2}. Below the screening threshold, participants with more trauma symptoms were flagged more often than their PHQ-8 predicted. Greater amounts of speech also predicted additional flags beyond PHQ-8. Specificity remains unresolved where depression and PTSD co-occur, including trauma-exposed and veteran populations. Ambient scribes now record entire clinical visits\cite{Tierney2024}, making the clinician's contribution to model inputs a practical concern beyond research interviews.

If the configuration is the instrument, changing a component changes the instrument. Changes in model precision and inference software, model size, instructions and transcription each altered substantial numbers of decisions. Five practices follow (Box~1). Report decision ranges across defensible configurations, extending multiverse and many-analyst approaches\cite{Steegen2016,Silberzahn2018,Botvinik2020} to clinical LLM reporting\cite{Gallifant2025}. Fix a neutral configuration and test rewording robustness. Calibrate against a validated criterion in a few dozen local patients, check transfer and report residual individual disagreement. Test specificity against neighbouring conditions. Freeze the complete stack, especially when following a patient over time. Otherwise, a future change in a patient's score could reflect a changed instrument as well as a changed condition. Longitudinal validation would need to distinguish these possibilities.

\begin{figure}[t]
\centering
\fbox{\begin{minipage}{0.95\textwidth}
\small
\textbf{Box 1 \textbar{} Five checks before a language-model rating informs care}

\par\smallskip\noindent\textbf{Report the range, not one number.} Rate with several defensible configurations and report how often their decisions disagree, including among configurations that pass checks of accuracy, sensitivity and specificity, and whether they disagree about how many people to flag or about which ones (Figs~\ref{fig:verdicts}, \ref{fig:decisions} and \ref{fig:whowhat}).
\par\smallskip\noindent\textbf{Fix a neutral configuration, and treat fragility as an alarm.} Choose the wording, answer format and scoring in advance, for example the questionnaire's own wording with digit answers. Reword the request: a rating that changes with the wording is fragile. Internal consistency did not rank configurations across models (Fig.~\ref{fig:reassurance}).
\par\smallskip\noindent\textbf{Calibrate locally, check transfer, and measure what remains.} A few dozen local patients who completed a validated questionnaire can set the threshold and give honest estimates of prevalence. Check whether a calibration carries over to a new setting rather than assuming it. A recalibration that preserves a configuration's ranking changes how many people are flagged, not whom it ranks highest, so measure the remaining disagreement about individuals (Figs~\ref{fig:decisions}c and \ref{fig:whowhat}).
\par\smallskip\noindent\textbf{Ask what else it reads.} Test specificity against neighbouring conditions such as PTSD with criteria that can separate them, and whether the clinician's words drive the rating (Fig.~\ref{fig:who}).
\par\smallskip\noindent\textbf{Freeze the whole stack.} Treat any change of model, precision, inference software, transcription or wording as a new instrument, and revalidate it (Fig.~\ref{fig:dials}).

\end{minipage}}
\end{figure}

Several limits define the scope. Corpus licences required local processing, restricting us to open models up to 21 billion parameters. We scored normalised answer probabilities after a fixed cue, without generated reasoning. Frontier models used as intended may differ. Size improved ranking and narrowed threshold variation without eliminating it; larger models may reduce these effects, but disappearance cannot be assumed. Our finite, deliberately broad grid makes inclusion consequential, motivating stricter sets, fixed configurations and single-choice comparisons. PHQ-8 was a pre-interview self-report screening criterion, not a diagnosis, although rater disagreement requires no criterion. No clinicians rated these transcripts, precluding direct comparison with human disagreement.

Both English-language sets followed one laboratory's protocol. Confirmation was small, with few women, and changed participants, interviewer and transcription, not clinical setting. Calibration transfer was therefore one related comparison. Analyses of how many versus who, recalibration, transfer, deferral and fairer label-free comparisons were added after confirmation and remain exploratory. Repeated splits of small cohorts are not independent samples. Finally, frozen code corrected termination for only some direct totals, and batching affected inference reproducibility. Both changed scores by fractions of a point in the re-scored sample (Methods).

Open questions extend beyond depression. Do frontier models escape the rater effect? Does disagreement persist across clinically equivalent paraphrases, and how does it compare with clinicians reading identical transcripts? How large are configuration effects relative to treatment-related change? Can raters be reconciled, or designed, to agree about individuals while remaining accurate against an independent clinical assessment? The same decomposition could inform assessments of essays, job applicants and online speech wherever a model's judgement becomes a measurement of a person.

A questionnaire makes its eight questions and cut-off visible, so readers can inspect and contest the terms on which it judges a person. A language-model rater can conceal equally consequential choices in software that appears merely to read the interview. Here, those choices shaped how many people were flagged and which people they were. Calibration made the raters more nearly right on average, yet left them divided about individuals. Clinical care meets people one at a time. When changing how we ask changes whom we identify, the instrument becomes part of the explanation. Responsible measurement makes that contribution visible, so that a judgement about a person can be understood and questioned by the person who must live with it.

\section*{Online Methods}

\paragraph{Data.} The discovery set was the Distress Analysis Interview Corpus -- Wizard of Oz (DAIC-WOZ)\cite{Gratch2014,DeVault2014}: 189 semi-structured interviews of adults, conducted by an animated virtual interviewer that a human operator controlled from another room, with manual transcripts of the interviewer's and the participant's turns. Before the interview, participants completed the PHQ-8, which scores eight symptoms over the preceding two weeks from 0 (not at all) to 3 (nearly every day), for a total of 0--24; a total of 10 or more is the standard screening threshold\cite{Kroenke2009}. A participant-level meta-analysis also supports this threshold for the related PHQ-9\cite{Levis2019}. A participant screened positive if their PHQ-8 total was 10 or more; this registered definition differs from the binary flag distributed with the corpora for one discovery and five confirmation participants (Supplementary Information). Totals were available for all 189 participants and item scores for 142 (complete for 141; the challenge training and development partitions\cite{Valstar2016}). The same participants' PTSD Checklist -- Civilian Version (PCL-C) severity was available from the extended corpus (E-DAIC)\cite{Ringeval2019}. The confirmation set comprised the 86 E-DAIC interviews (sessions 600--718) conducted by a fully autonomous version of the virtual interviewer, with automatic speech recognition (ASR) transcripts of the participant's speech and PHQ-8 and PCL-C scores (PCL-C missing for two). In the discovery set, 30.2\% of participants screened positive (57 of 189; 87 women and 102 men); in the confirmation set, 33.7\% (29 of 86; 18 women and 68 men). The corpora were used under their end-user licence agreements; all processing ran on a local workstation, and no transcript text was sent to any external service.

\paragraph{Models.} We used 11 open-weight instruction-tuned models from six developers, chosen because their weights were already on the workstation: Qwen2.5-Instruct with 1.5, 3, 7 and 14 billion parameters\cite{Qwen2025}; Llama-3.2-Instruct 1B and 3B and Llama-3.1-8B-Instruct\cite{Llama3}; Mistral-7B-Instruct-v0.3\cite{Mistral2023}; Phi-3.5-mini-instruct\cite{Phi3}; DeepSeek-R1-Distill-Qwen-7B\cite{DeepSeekR1}; and gpt-oss-20b\cite{GptOss2025}. Qwen2.5 1.5B, 3B and 7B ran in bfloat16 with the Transformers library\cite{Wolf2020} on the Apple GPU; the others ran as 8-bit (Llama-3.2-1B) or 4-bit quantized GGUF files with llama.cpp. For the reasoning model the reasoning block was left empty, and gpt-oss-20b answered in its final channel at low reasoning effort. As a check on the deployment stack, Qwen2.5-7B was also run as a 4-bit GGUF file with llama.cpp.

\paragraph{Prompts.} Each prompt, sent in the model's own chat format, began with a one-sentence description of the transcript and the transcript itself, followed by a task instruction (wording), a description of the response options (format) and one PHQ-8 item. The five wordings represented distinct instructions: \emph{instrument}, complete the PHQ-8 on the participant's behalf using the interview as the only source; \emph{clinician}, an experienced psychiatrist rating the participant; \emph{evidence}, rate only symptoms the participant explicitly reports and rate undiscussed symptoms as absent; \emph{role-play}, answer as the participant would about themselves; and \emph{inferential}, use every cue, including tone, word choice and avoidance. The evidence, role-play and inferential wordings change the assessment policy as well as the words; the clinician wording changes only who is said to rate. The three formats were digits 0--3, letters A--D with A meaning `not at all', and letters A--D with A meaning `nearly every day', each with the PHQ-8 anchor labels. Items used the PHQ-8 wording with the standard two-week stem. For the direct total, the wording was followed by a request for the participant's total PHQ-8 score from 0 to 24, answered with digits. Exact texts are given in Supplementary Information and the code.

\paragraph{Reading the answer.} We did not sample. The assistant's reply was prefilled with an answer cue (``Answer:'' for items, ``Score:'' for totals), and each answer option was scored by its probability as the next tokens, summed over tokenisation variants (with and without a leading space) and renormalised over the options. The unnormalised total was retained as a measure of format compliance. This is normalised candidate-label likelihood scoring. The expected-value rule uses the whole distribution, and the most-likely-option rule takes the option with the highest normalised probability, which need not coincide with the answer that free or token-by-token greedy generation would produce. For totals, we scored the 25 numbers 0--24 as token sequences. The intended rule multiplies each number's probability by the probability that it ends where it does (the next token is not a digit); the frozen code applied this factor only to numbers whose tokens begin a longer candidate. Re-scoring the direct totals of \TotParts{} participants spread over the PHQ-8 range with both rules from the same forward pass, for \TotModels{} models and the participant view, changed expected totals by at most \TotDeltaMax{} points in that sample and no screening decision (Supplementary Information); this measures the defect in a sample rather than repairing it, and item sums are unaffected. All prompts for one transcript share the transcript as a prefix; we encoded it once and evaluated all continuations in a few forward passes with tree-structured attention, which reproduced one-prompt-at-a-time evaluation to $10^{-4}$ in log-probability in 32-bit arithmetic for the prompts tested. Inference was not bit-for-bit reproducible across batch compositions: re-running the same direct-total prompts reproduced the saved expected totals to a median of \BatchMed{} points (maximum \BatchMax{}), changing \BatchFlips{} of \BatchRows{} decisions.

\paragraph{Raters and scores.} A rater (configuration) is a combination of model (11), transcript view (participant speech only; full dialogue), and either a wording, a format and a scoring rule applied to the eight items, summed (5 $\times$ 3 $\times$ 2), or a wording and a scoring rule applied to the direct total (5 $\times$ 2): $11 \times 2 \times 40 = 880$ raters. The expected-value rule scores an item as $\sum_k k\,p_k$ and the mode rule as $\arg\max_k p_k$. A rater screens a participant positive if its score is 10 or more. Two further views served as checks: the interviewer's speech only, which contains no report from the participant but is known to leak depression status through the interviewer's follow-up questions\cite{Burdisso2024}, and ASR transcripts of the same participants' speech.

\paragraph{Quality checks.} Two secondary outputs of the first run, for Qwen2.5 1.5B (the interviewer's speech for one participant and the automatic transcript for another), had no finite probabilities, and the frozen analysis's median over models silently omitted that model. We regenerated both with the frozen code, which gave finite outputs with nearly all probability on the answer options, kept the original files, and re-ran the frozen analysis: primary results were unchanged, and the prediction about the interviewer's speech, previously met, was no longer supported. A check of every raw output against an explicit manifest of the expected models, conditions and participants now runs before the frozen analysis. The only permitted exclusions are three participants whose interviewer transcripts were empty. The check also verifies the registered dimensions and that the stored option probabilities sum to one within their rounding. It rejects empty, truncated or incomplete files and flags the two original outputs. In six discovery transcripts, empty turns had been converted to the literal token `nan' (1--12 tokens each); excluding these participants changed the median AUC from 0.753 to 0.757.

\paragraph{Estimands.} For each rater we computed the area under the ROC curve (AUC) for PHQ-8 $\geq 10$\cite{Hanley1982}, Spearman's correlation with the PHQ-8 total, the mean absolute error and bias (mean over-rating), and the share screened positive. For each participant, $\pi_i$ is the share of raters that screen them positive; a participant is discordant if $0<\pi_i<1$ and contested if $0.2\leq\pi_i\leq0.8$. The chance that two different raters drawn at random from $K$ disagree about a participant is $2\pi_i(1-\pi_i)K/(K-1)$; we report its mean over participants (the frozen regularities used $2\pi_i(1-\pi_i)$, which differs negligibly). The expected share of participants receiving both decisions from $k$ raters drawn at random without replacement is computed exactly, $1-[\binom{m_i}{k}+\binom{K-m_i}{k}]/\binom{K}{k}$ averaged over participants, with $m_i$ the number of raters screening participant $i$ positive. For two raters with positive rates $p_A$ and $p_B$, $P(A\neq B) = |p_A-p_B| + 2\min\{P(A{=}1,B{=}0), P(A{=}0,B{=}1)\}$, an algebraic identity that separates the least disagreement forced by unequal rates from the rest, without assigning causes. At equal capacity, each rater selects its $q$ highest-scoring participants, with $q$ the number of participants screening positive by self-report unless stated; expected-value scores had no ties at the boundaries used. We compared the between-rater variance of the AUC with its participant-bootstrap sampling variance. The generalizability study treated person, model, wording, format and view as crossed random facets of the expected item sum (330 raters in discovery), with variance components estimated from expected mean squares\cite{Cronbach1972,Brennan2001}. We report the person main effect and, separately, all components involving the person. Decision studies give the generalizability ($E\rho^2$) and dependability ($\Phi$) coefficients when scores are averaged over several models, wordings and formats; in discovery they also average both views unless stated. For each item-sum rater we computed Cronbach's $\alpha$ across the eight items; the correlation between two independent runs sampled at temperature 1, which follows analytically from the answer distributions; and the mean correlation with the same rater under the other wordings. We related each index to the AUC across raters and within models. For group inference, we computed each rater's standardised female--male difference (the difference in means divided by the square root of the mean of the two group variances) and compared its sign and size with the difference in the criterion\cite{Gelman2014}. Convergent and discriminant validity compared each rater's correlation with the PHQ-8 and with PTSD severity by the Meng--Rosenthal--Rubin test for correlated correlations\cite{Meng1992}; Steiger's pooled form\cite{Steiger1980} and a paired participant bootstrap of the difference gave the same conclusions. Agreement between transcript sources and between deployment stacks used decision flips and ICC(2,1)\cite{Shrout1979}. Pre-registered remedies were averaging across raters or models; isotonic calibration of each rater's total score to the PHQ-8 total with 20 or 40 labelled participants (200 random draws, evaluated on the rest), which calibrates scores rather than probabilities; and prediction-powered estimation of prevalence with the same labels\cite{Angelopoulos2023}.

\paragraph{Further analyses.} Sensitivity, specificity and the share screened positive follow from each rater's decisions. The share screened positive was related to the rater's bias by a normal-ogive curve, $\Phi((b-c)/s)$, fitted by least squares; the number of additional people flagged per point of bias at the centre is $100\,\phi(0)/s$. A model's typical configuration is the one with its median AUC. Effects of wording and letter order compare model medians across the corresponding configurations; effects of aggregation compare item sums and direct totals with digit answers only, because direct totals were asked only with digits (the frozen regularity, Extended Data Table~\ref{tab:rules}, counted item sums in all formats and found 7 of 11 models); and effects of size use the log parameter count. Partial Spearman correlations remove a linear trend in the ranks of the PHQ-8 from both variables and are tested with $n-3$ degrees of freedom. Decision changes between neighbouring model sizes compare identical configurations. The label-free rule chose, for each model, the expected-value item-sum rater with the highest agreement across wordings (ties broken by sampling reliability), and was compared, as specified, with the median AUC of the model's item-sum raters. These analyses were exploratory in the discovery set. The label-free rule, with the prediction that it would beat the model median in at least 9 of 11 models, and nine regularities, each with a pass criterion, were specified in time-stamped addenda after the discovery analysis and before any confirmation output was compared with the criterion, while confirmation outputs were still being generated; all are reported. The criteria are descriptive, and because the regularities share participants and raters they are not independent replications.

\paragraph{Analyses added after the confirmation analysis.} The following analyses were not pre-specified and are exploratory in both sets. Pairwise disagreement in stricter sets of raters (AUC $\geq$ 0.70 and 5--95\% screened positive; AUC, sensitivity and specificity all $\geq$ 0.70), and the exact expected discordance among $k$ raters. The identity for the original decisions, and equal-capacity selection at 10--50\% of participants, for expected-value raters with AUC $\geq$ 0.70 (a selection that uses the whole cohort's labels), all expected-value raters and one fixed configuration per model (instrument wording, digits, participant's speech, expected value), with each rater's disagreement with the criterion at the self-reported prevalence. Single-choice comparisons over all pairs of configurations identical except in one choice. The model-to-person variance ratio with digit answers only, for the instrument wording with digits and the participant's speech, and with each developer's models left out. The label-free rule against the median of expected-value item sums, the median of those with digit answers, and the fixed neutral configuration. Recalibration on held-out participants: in each of 200 random splits, 20, 40 or 80 participants (20 or 40 in confirmation) formed a labelled sample and the rest were held out; three rater sets were recalibrated, namely raters with AUC $\geq$ 0.70 in the labelled sample, one fixed configuration per model, and, for comparison, raters with AUC $\geq$ 0.70 in the whole cohort, whose admission uses the held-out labels; the rules were subtracting the mean over-rating, linear regression of the PHQ-8 total on the score, logistic regression for PHQ-8 $\geq$ 10 with a weak ridge penalty (deciding at a probability of 0.5), the attainable threshold whose labelled positive count is closest to the number of labelled positives, compared exactly as integers with ties going to the higher threshold (ties among scores can make the target unattainable; achieved counts are reported, a miss of 5 percentage points counting as within 5), and the highest threshold that keeps at least 80\% of the labelled positives (the exact order statistic). We report held-out accuracy, sensitivity, specificity, pairwise disagreement, the median absolute error of the share flagged with its 2.5th and 97.5th percentiles across splits, and achieved training targets. A defensible core (instrument or clinician wording, digits, expected values, symptom sums; every view) was analysed alongside these sets without any selection by outcome. Deferral was compared at matched coverage on the same held-out participants: a single recalibrated rater, the majority of raters, the majority after setting aside participants on whom fewer than a given share of raters agreed, single raters on the same retained participants, and each single rater setting aside its own closest calls, those nearest its cut-off, to reach the same coverage; retained participants with and without a positive screen are counted. Where disagreement falls was summarised by self-reported PHQ-8 band, at the threshold and at equal capacity. Three headline estimands were bootstrapped over participants with the rater sets held fixed. Transfer carried each configuration's discovery offset or discovery thresholds (matching the discovery prevalence, or keeping 80\% of discovery positives), for configurations with AUC $\geq$ 0.70 in discovery and for the single most accurate discovery configuration, unchanged to confirmation. Prevalence intervals from 40 labels (2,000 draws): prediction-powered with the registered formula, prediction-powered for the finite cohort, and from the labels alone. The discovery normal-ogive curve applied unchanged to confirmation. The decision rules and their tie handling are implemented and tested in the analysis code.

\paragraph{Pre-registration and confirmation.} The study was pre-registered on OSF Registries (\url{https://osf.io/8qg2m/}) before any model output was compared with the criterion, and the scoring and analysis code were fixed at the same time. The pre-registered hypotheses were that the validity, individual screening decisions and implied prevalence of the ratings depend substantially on configuration choices, and that indices of rating consistency do not indicate which configurations are valid. Model outputs for the confirmation set were generated after the discovery analysis was complete, with the locked code, and analysed once. Deviations from the plan, including the regenerated outputs and the scoring defect, are listed in Supplementary Information.

\paragraph{Statistics.} Intervals for AUCs and for the gain of the label-free rule are percentile 95\% intervals from 2,000 bootstrap resamples of participants. The pre-registered calibration and prediction-powered analyses used 200 random draws of labelled participants (50 for the display in Extended Data Fig.~\ref{fig:ed-remedies}b); the exploratory analyses used 200 splits for recalibration and deferral, 2,000 draws for prevalence intervals, 1,000 paired bootstrap resamples for the specificity check and 1,000 participant resamples, with the rater sets held fixed, for the headline intervals, which are therefore conditional on the raters. Split-to-split ranges describe variation over repeated splits of one cohort and are not confidence intervals. All tests are two-sided unless stated.

\section*{Data availability}
DAIC-WOZ and E-DAIC are available from the University of Southern California Institute for Creative Technologies under a data use agreement. Only aggregate results are distributed with the code.

\section*{Code availability}
The scoring and analysis code will be made available on publication at a public repository.

\section*{Acknowledgements}
This work was supported in part by a 2025 NARSAD Young Investigator Grant from the Brain \& Behavior Research Foundation, funded by The Sidney R. Baer, Jr. Foundation (Grant ID 33941). B.L. is also supported by the U.S. National Institutes of Health (NIH), the U.S. Department of Veterans Affairs (VA), the Hasso Plattner Foundation, and the Windreich Family Foundation. We thank the University of Southern California Institute for Creative Technologies for making the interview corpora available.

\section*{Author contributions}
B.L. conceived the study, performed the analyses, and wrote the manuscript.

\section*{Competing interests}
The author declares no competing interests.

\clearpage
\section*{Extended Data}
\renewcommand{\theHfigure}{ed.\arabic{figure}}
\setcounter{figure}{0}
\renewcommand{\figurename}{Extended Data Fig.}
\renewcommand{\theHtable}{ed.\arabic{table}}
\setcounter{table}{0}
\renewcommand{\tablename}{Extended Data Table}

\begin{table}[h]
\small
\caption{\textbf{Regularities found in the discovery set and checked in new interviews.} Discovery values are exploratory; the criteria were specified before the confirmation analysis. $\pi$, a participant's share of raters with AUC $\geq$ 0.70 screening them positive. For the first regularity, confirmation values refit the curve; applied unchanged, the discovery curve gives $R^2$ = \LockedRtwo{}. $P$ values for partial correlations use $n-3$ degrees of freedom ($P$ = \RulePtsdPC{} for the PTSD regularity in confirmation). With digit answers only, item sums over-rate in \RuleItemSumMatched{} of 11 models in both sets. The regularities share participants and raters, so they are not independent tests.}
\label{tab:rules}

\begin{tabularx}{\textwidth}{@{}Xllll@{}}
\toprule
Regularity & Criterion & Discovery & Confirmation & Held\\
\midrule
The share screened positive follows a configuration's mean over-rating (probit $R^2$; referrals per 100 per point) & $R^2 \geq$ 0.75 & 0.91; 13 & 0.91; 14 & yes\\
Configurations with the same AUC ($\pm$0.02) differ in share screened positive (median, points) & $\geq$ 15 & 28 & 28 & yes\\
Participants reporting few symptoms (PHQ-8 $\leq$ 4) are screened positive (mean share of configurations with AUC $\geq$ 0.70) & $\geq$ 20\% & 35\% & 44\% & yes\\
Larger models discriminate better (Spearman, size vs median AUC), yet no model's ratings correlate with the PHQ-8 more than 0.10 above their correlation with PTSD severity (median difference; not a test of diagnostic separation) & $>$ 0.5; none $>$ 0.10 & 0.82; max 0.04 & 0.77; max 0.01 & yes\\
Below the threshold, more PTSD symptoms go with a larger share screened positive at equal PHQ-8 (partial Spearman) & $>$ 0, $P <$ 0.05 & 0.36 & 0.23 & yes\\
Rating only explicitly reported symptoms lowers referrals; reversing the answer letters moves them (median points) & 8 and 6 of 11 models & 27; 21 & 31; 24 & yes\\
Summing item ratings over-rates relative to rating the total directly (models; as frozen, item sums in all formats) & $\geq$ 6 of 11 & 7 of 11 & 7 of 11 & yes\\
Talkative participants are screened positive more often at equal PHQ-8 (partial Spearman) & $>$ 0, $P <$ 0.05 & 0.25 & 0.26 & yes\\
The most accurate model's own configurations disagree about participants (mean $2\pi(1-\pi)$, as frozen) & $\geq$ 0.15 & 0.24 & 0.26 & yes\\
\bottomrule
\end{tabularx}

\end{table}

\begin{figure}[h]
  \centering
  \includegraphics[width=\textwidth]{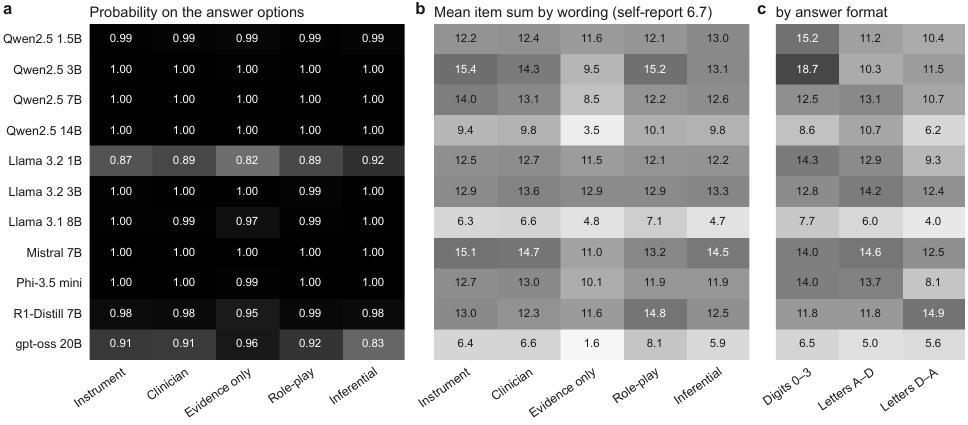}
  \caption{\textbf{Format compliance and rated severity.} \textbf{a}, Median probability that each model placed on the answer options, by wording (participant view). \textbf{b}, Mean expected item sum by model and wording; the self-reported mean PHQ-8 is given in the panel title. \textbf{c}, Mean expected item sum by model and answer format.}
  \label{fig:ed-compliance}
\end{figure}

\begin{figure}[h]
  \centering
  \includegraphics[width=0.62\textwidth]{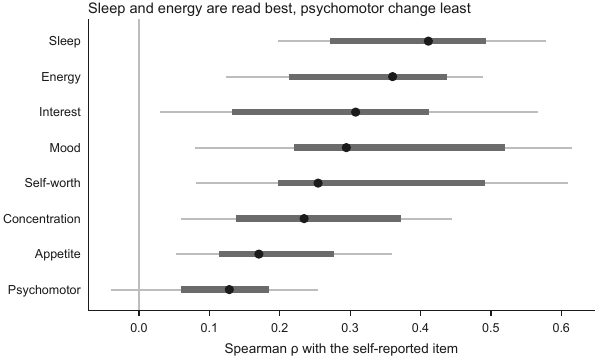}
  \caption{\textbf{Agreement with individual symptoms.} Spearman correlation between each model's expected item rating and the participant's self-reported score on the same item, for the 141 participants with complete item scores (participant view), per symptom: dot, median model; thick line, interquartile range over models; thin line, range over models; each model's value is its median over wordings and formats. Exploratory analysis.}
  \label{fig:ed-items}
\end{figure}

\begin{figure}[h]
  \centering
  \includegraphics[width=\textwidth]{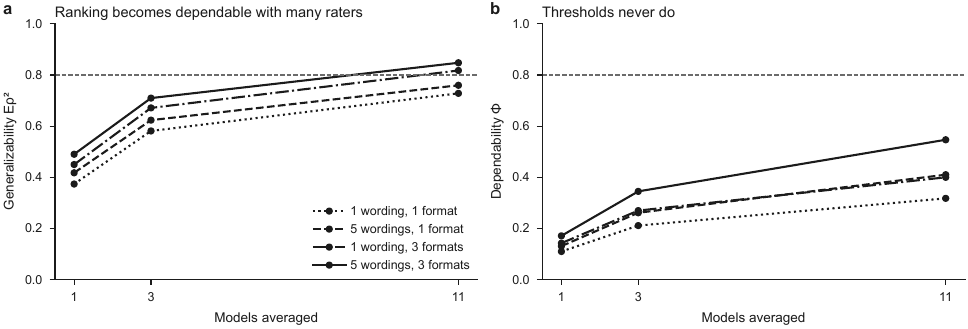}
  \caption{\textbf{Decision studies.} Generalizability ($E\rho^2$; \textbf{a}) and dependability ($\Phi$; \textbf{b}) of the expected item sum (discovery), when scores are averaged over the given number of models and over one or all wordings and formats, with both views of the conversation averaged. Dashed line, 0.80.}
  \label{fig:ed-dstudy}
\end{figure}

\begin{figure}[h]
  \centering
  \includegraphics[width=\textwidth]{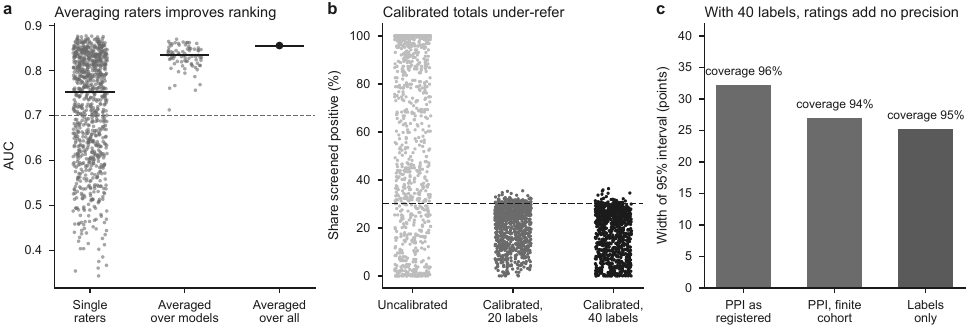}
  \caption{\textbf{What averaging and labels do} (discovery). \textbf{a}, AUC of single raters, of scores averaged over models for each configuration of the other choices, and of the average over all raters; lines, medians; dashed line, 0.70. \textbf{b}, Share screened positive by each rater before and after isotonic calibration of its total score against 20 or 40 labelled participants (mean over 50 draws); dashed line, self-reported prevalence. \textbf{c}, Mean width of 95\% intervals for the prevalence from 40 labelled participants (2,000 draws): prediction-powered intervals with the registered formula and for the finite cohort, and intervals from the labels alone; observed coverage above the bars.}
  \label{fig:ed-remedies}
\end{figure}

\begin{figure}[h]
  \centering
  \includegraphics[width=\textwidth]{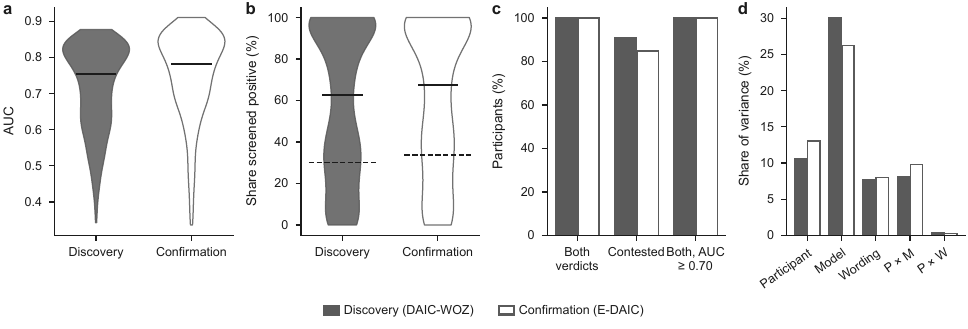}
  \caption{\textbf{Discovery and confirmation.} Discovery (DAIC-WOZ, dark grey) and confirmation (E-DAIC sessions 600--718, automatic transcripts, white). \textbf{a}, AUC of every rater; lines, medians. \textbf{b}, Share screened positive; dashed lines, self-reported prevalence. \textbf{c}, Participants receiving both decisions, contested decisions (20--80\% of raters), and both decisions among raters with AUC $\geq$ 0.70. \textbf{d}, Shares of the variance of the item-sum score: participant, model, wording, participant $\times$ model (P $\times$ M) and participant $\times$ wording (P $\times$ W).}
  \label{fig:ed-confirm}
\end{figure}

\begin{figure}[h]
  \centering
  \includegraphics[width=\textwidth]{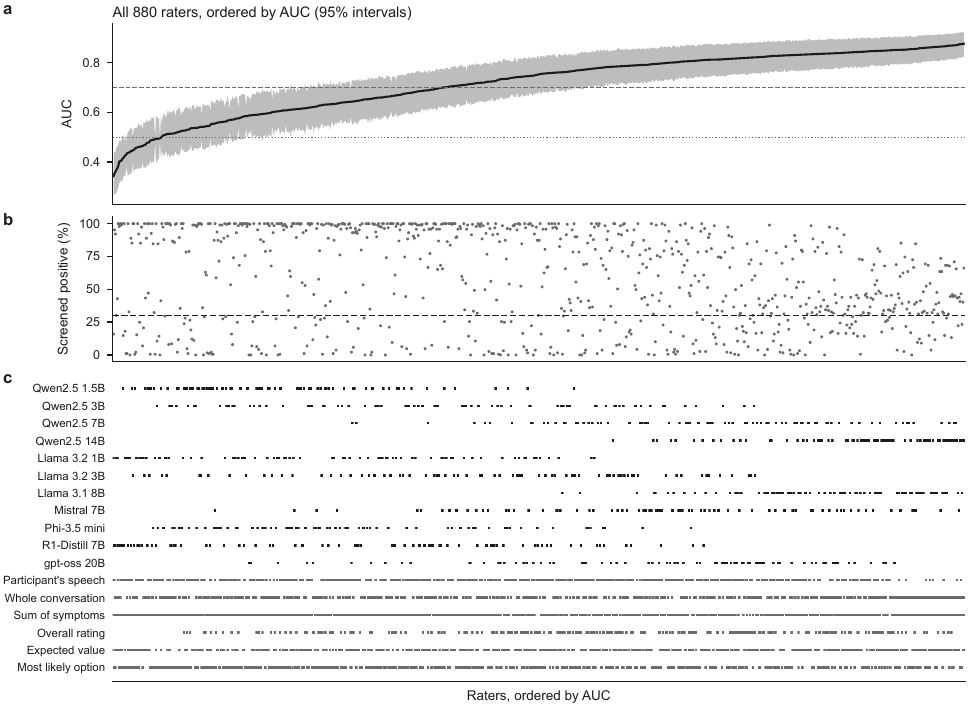}
  \caption{\textbf{All raters, ordered by accuracy} (discovery). \textbf{a}, AUC of every rater, ordered, with 95\% bootstrap intervals; dotted line, chance; dashed line, 0.70. \textbf{b}, Share screened positive by the same raters; dashed line, self-reported prevalence. \textbf{c}, The choices that make up each rater.}
  \label{fig:ed-multiverse}
\end{figure}

\begin{figure}[h]
  \centering
  \includegraphics[width=\textwidth]{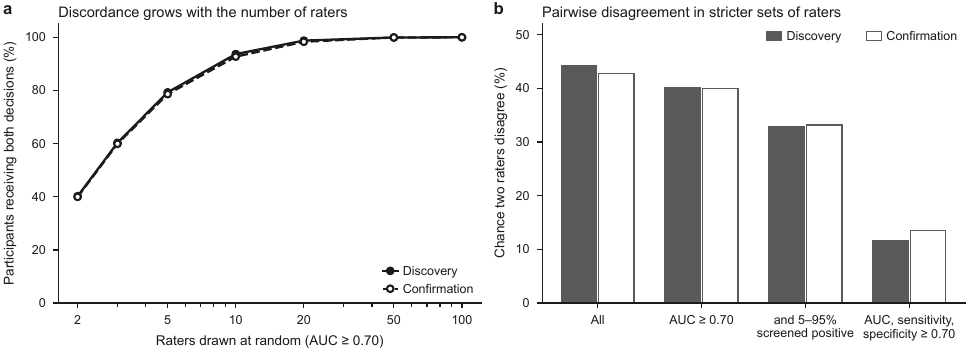}
  \caption{\textbf{How disagreement depends on which raters are counted.} \textbf{a}, Expected share of participants who receive both decisions from $k$ raters with AUC $\geq$ 0.70 drawn at random without replacement (exact), against $k$; filled markers and solid line, discovery; open markers and dashed line, confirmation. \textbf{b}, Chance that two different raters disagree about a participant, among all raters, raters with AUC $\geq$ 0.70, those that also screen 5--95\% of participants positive, and those with AUC, sensitivity and specificity all $\geq$ 0.70; dark grey, discovery; white, confirmation. Exploratory analyses.}
  \label{fig:ed-grid}
\end{figure}

\begin{figure}[h]
  \centering
  \includegraphics[width=\textwidth]{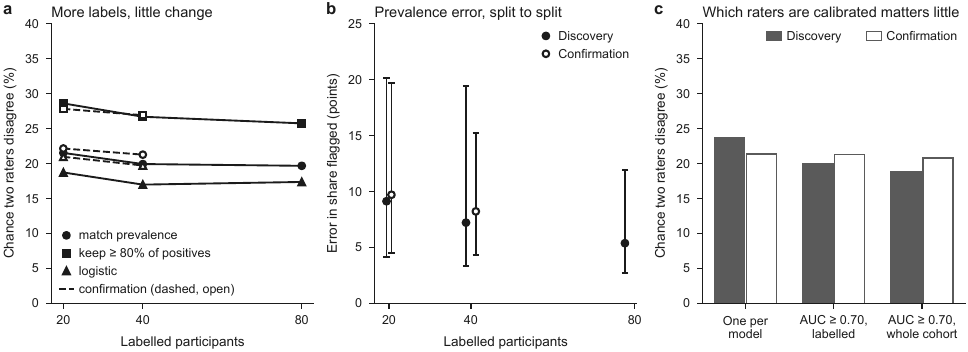}
  \caption{\textbf{Recalibration in detail.} \textbf{a}, The chance that two recalibrated raters disagree about a held-out participant against the number of labelled participants, for raters admitted by their AUC in the labelled sample and recalibrated to match the labelled prevalence, to keep at least 80\% of labelled positives, or by logistic regression; solid and filled, discovery; dashed and open, confirmation. \textbf{b}, Median absolute error of the share flagged with the prevalence-matched threshold; bars, 2.5th to 97.5th percentiles across splits (not confidence intervals). \textbf{c}, Disagreement after matching the prevalence with 40 labels for three rater sets: one fixed configuration per model, raters admitted by their AUC in the labelled sample, and raters admitted by their AUC in the whole cohort, a selection that uses the held-out labels. Means over 200 splits; exploratory analyses.}
  \label{fig:ed-recal}
\end{figure}

\begin{figure}[h]
  \centering
  \includegraphics[width=\textwidth]{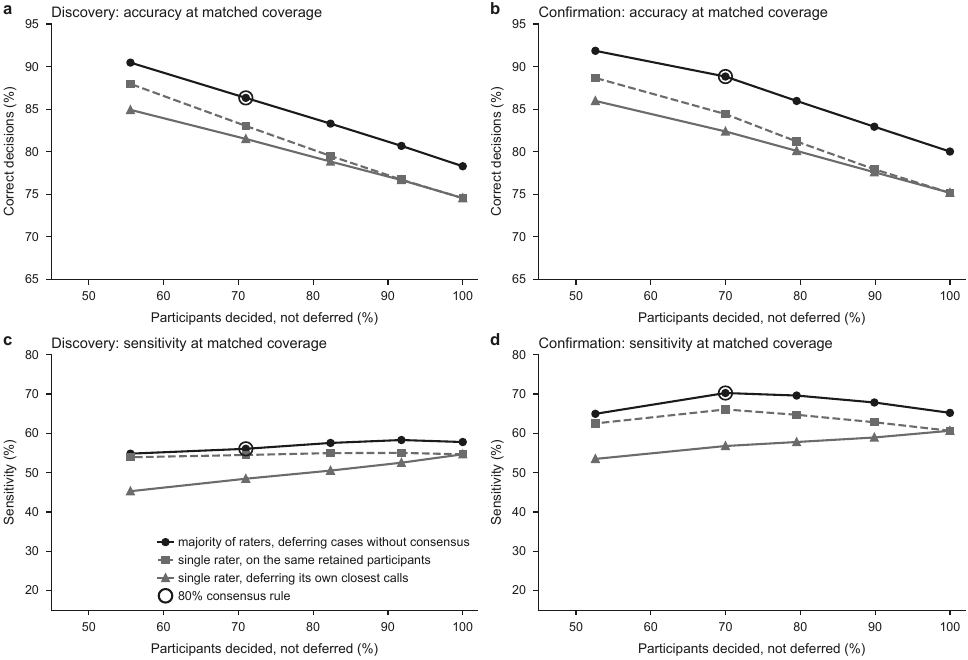}
  \caption{\textbf{Deferral at matched coverage.} For raters admitted by their AUC on 40 labelled participants and recalibrated to the labelled prevalence, accuracy (\textbf{a}, discovery; \textbf{b}, confirmation) and sensitivity (\textbf{c}, discovery; \textbf{d}, confirmation) on the held-out participants who are decided rather than deferred, against the share decided: the majority of raters for all participants (50\% consensus) and after setting aside participants on whom fewer than 60, 70, 80 or 90\% of raters agree (black; unanimous consensus is not shown because it left fewer than 3\% of participants); single raters on the same retained participants (grey, dashed); and single raters setting aside their own closest calls to reach the same coverage (grey, solid). Large circle, the 80\% consensus rule. Means over 200 splits; exploratory analyses.}
  \label{fig:ed-defer}
\end{figure}

\clearpage
\section*{Supplementary Information}
\setcounter{table}{0}
\renewcommand{\tablename}{Supplementary Table}
\renewcommand{\theHtable}{si.\arabic{table}}
\renewcommand{\theHsection}{si.\arabic{section}}
\section{Supplementary Methods}

\subsection{Prompts}
Each prompt was a single user turn in the model's own chat format. Its components appeared in a fixed order: a one-sentence description of the transcript view; the transcript, enclosed in \texttt{<transcript>} tags, with turns labelled ``Interviewer:'' and ``Participant:'' in the full-dialogue view, ``Interviewer:'' in the interviewer view and unlabelled in the participant view; the template's task instruction; the response format; and one PHQ-8 item preceded by the stem ``Over the last 2 weeks, how often has the participant been bothered by the following problem?''. The assistant's reply began with the answer cue specified by that format. For the direct total, the format and item were replaced by ``Considering all eight PHQ-8 problems together, what is the participant's total PHQ-8 score (from 0 to 24)? Reply in the form `Score: N'.'', and the reply was prefilled with ``Score:''. Supplementary Table~\ref{tab:prompts} gives every text.

\begin{table}[H]
\small
\caption{Prompt texts. The transcript follows the first line; the template, the format and the item follow the transcript.}
\label{tab:prompts}
\begin{tabularx}{\textwidth}{@{}lX@{}}
\toprule
\multicolumn{2}{@{}l}{\textbf{Transcript view (first line of every prompt)}}\\
participant & Below is a transcript of a clinical research interview. Only the participant's speech is shown.\\
full & Below is a transcript of a clinical research interview between an interviewer and a participant.\\
interviewer & Below is a transcript of a clinical research interview. Only the interviewer's speech is shown.\\
\midrule
\multicolumn{2}{@{}l}{\textbf{Template}}\\
instrument & Task: complete the PHQ-8 depression questionnaire on behalf of the participant, using the interview as your only source of information.\\
clinician & You are an experienced psychiatrist. Using your clinical judgement about this interview, rate the participant on the PHQ-8 depression scale.\\
evidence & Rate each PHQ-8 symptom only from what the participant explicitly reports in the interview. If a symptom was not discussed, rate it as 'Not at all'. Do not speculate.\\
role-play & Imagine that you are the participant in this interview. Answer the PHQ-8 questionnaire the way they would answer it about themselves, given everything they said.\\
inferential & Use every available cue, including indirect ones such as tone, word choice and what the participant avoids talking about, to infer how the participant would score on the PHQ-8.\\
\midrule
\multicolumn{2}{@{}l}{\textbf{Response format (answer cue prefilled in the reply)}}\\
digits & Answer each question with a single digit: 0 = Not at all, 1 = Several days, 2 = More than half the days, 3 = Nearly every day. Reply with the digit only. Cue: ``Answer:''\\
letters & Answer each question with a single letter: A = Not at all, B = Several days, C = More than half the days, D = Nearly every day. Reply with the letter only. Cue: ``Answer:''\\
letters reversed & Answer each question with a single letter: A = Nearly every day, B = More than half the days, C = Several days, D = Not at all. Reply with the letter only. Cue: ``Answer:''\\
\midrule
\multicolumn{2}{@{}l}{\textbf{Item stem and items}}\\
stem & Over the last 2 weeks, how often has the participant been bothered by the following problem?\\
item 1 & Little interest or pleasure in doing things.\\
item 2 & Feeling down, depressed, or hopeless.\\
item 3 & Trouble falling or staying asleep, or sleeping too much.\\
item 4 & Feeling tired or having little energy.\\
item 5 & Poor appetite or overeating.\\
item 6 & Feeling bad about yourself, or that you are a failure or have let yourself or your family down.\\
item 7 & Trouble concentrating on things, such as reading the newspaper or watching television.\\
item 8 & Moving or speaking so slowly that other people could have noticed, or the opposite, being so fidgety or restless that you have been moving around a lot more than usual.\\
\midrule
direct total & Considering all eight PHQ-8 problems together, what is the participant's total PHQ-8 score (from 0 to 24)? Reply in the form 'Score: N'. Cue: ``Score:''\\
\bottomrule
\end{tabularx}

\end{table}

\subsection{Models}
Supplementary Table~\ref{tab:models} lists the models, their runtimes and the exact weights. Transformers models ran on the Apple GPU (Metal Performance Shaders) in bfloat16; GGUF files ran with llama.cpp (llama-cpp-python 0.3.16) with all layers on the GPU. The reasoning model received an empty reasoning block, and gpt-oss-20b was set to low reasoning effort and read in its final channel, so that every model answered directly.

\begin{table}[H]
\small
\caption{Models and weights. $^{\dagger}$Numerical-precision check only; not part of the multiverse.}
\label{tab:models}

\begin{tabularx}{\textwidth}{@{}lllp{3.2cm}X@{}}
\toprule
Model & Developer & Parameters (B) & Runtime and precision & Weights\\
\midrule
Qwen2.5 1.5B & Alibaba & 1.5 & Transformers, bf16 & \texttt{Qwen/Qwen2.5-\allowbreak{}1.5B-\allowbreak{}Instruct (revision 989aa79)}\\
Qwen2.5 3B & Alibaba & 3.1 & Transformers, bf16 & \texttt{Qwen/Qwen2.5-\allowbreak{}3B-\allowbreak{}Instruct (revision aa8e725)}\\
Qwen2.5 7B & Alibaba & 7.6 & Transformers, bf16 & \texttt{Qwen/Qwen2.5-\allowbreak{}7B-\allowbreak{}Instruct (revision a09a354)}\\
Qwen2.5 14B & Alibaba & 14.8 & llama.cpp, Q4\_K\_M & \texttt{Ollama library qwen2.5:14b; SHA-\allowbreak{}256 2049f5674b1e}\\
R1-Distill 7B & DeepSeek & 7.6 & llama.cpp, Q4\_K\_M & \texttt{DeepSeek-\allowbreak{}R1-\allowbreak{}Distill-\allowbreak{}Qwen-\allowbreak{}7B-\allowbreak{}Q4\_\allowbreak{}K\_\allowbreak{}M.gguf; SHA-\allowbreak{}256 731ece8d06dc}\\
Llama 3.2 1B & Meta & 1.2 & llama.cpp, Q8\_0 & \texttt{Ollama library llama3.2:1b; SHA-\allowbreak{}256 74701a8c35f6}\\
Llama 3.2 3B & Meta & 3.2 & llama.cpp, Q4\_K\_M & \texttt{llama-\allowbreak{}3.2-\allowbreak{}3b-\allowbreak{}instruct.Q4\_\allowbreak{}K\_\allowbreak{}M.gguf; SHA-\allowbreak{}256 6c1a2b411610}\\
Llama 3.1 8B & Meta & 8 & llama.cpp, Q4\_K\_M & \texttt{Ollama library llama3.1:8b; SHA-\allowbreak{}256 667b0c1932bc}\\
Phi-3.5 mini & Microsoft & 3.8 & llama.cpp, Q4\_0 & \texttt{Ollama library phi3.5:latest; SHA-\allowbreak{}256 b5374915da53}\\
Mistral 7B & Mistral AI & 7.2 & llama.cpp, Q4\_K\_M & \texttt{mistral-\allowbreak{}7b-\allowbreak{}instruct-\allowbreak{}v0.3.Q4\_\allowbreak{}K\_\allowbreak{}M.gguf; SHA-\allowbreak{}256 1270d22c0fbb}\\
gpt-oss 20B & OpenAI & 21 & llama.cpp, MXFP4/Q4\_K\_M & \texttt{gpt-\allowbreak{}oss-\allowbreak{}20b-\allowbreak{}Q4\_\allowbreak{}K\_\allowbreak{}M.gguf; SHA-\allowbreak{}256 c27536640e41}\\
Qwen2.5 7B (4-bit)$^{\dagger}$ & Alibaba & 7.6 & llama.cpp, Q4\_K\_M & \texttt{Ollama library qwen2.5:7b; SHA-\allowbreak{}256 2bada8a74506}\\
\bottomrule
\end{tabularx}

\end{table}

\subsection{Reading the answer}
For each item we read the probability of every response option at the position after the answer cue, summing the probabilities of tokenisation variants (with and without a leading space). The four probabilities were renormalised to sum to one; their unnormalised sum (the option mass) measures how completely the model complied with the format. For the direct total, each number from 0 to 24 was scored as a token sequence. The intended rule multiplies its probability by the probability that the next token is not a digit, so that, for example, ``1'' is not counted when the model goes on to write ``12''. The frozen code applied this factor only to numbers whose token sequence begins a longer candidate. Supplementary Table~\ref{tab:rescore} gives the size of the difference: we re-scored the direct totals of 25 participants spread over the PHQ-8 range for four models, assembling both rules from the same forward pass. The 25 probabilities were renormalised in the same way. The expected-value rule scores an item as $\sum_k k\,p_k$ and the mode rule as $\arg\max_k p_k$. This is normalised candidate-label likelihood scoring: each option's probability sums its tokenisation variants, and the mode is the option with the highest normalised probability. It need not coincide with the answer of free or token-by-token greedy generation; for example, an option whose two token variants each have probability 0.25 is chosen over an option with a single variant of probability 0.40, whereas greedy generation would emit the latter.

\begin{table}[H]
\small
\centering
\caption{Direct totals re-scored for 25 participants (five wordings each). Termination rule: the factor applied to every candidate against the frozen rule, from the same forward pass; flips, screening decisions that change under either scoring rule. Re-run against saved: the frozen rule in the new pass against the saved outputs (different batch composition).}
\label{tab:rescore}

\begin{tabular}{@{}lrrrrrr@{}}
\toprule
 & & \multicolumn{2}{l}{Termination rule} & \multicolumn{3}{l}{Re-run against saved}\\
Model & Totals & max $|\Delta|$ & Flips & median $|\Delta|$ & max $|\Delta|$ & Flips\\
\midrule
gpt-oss 20B & 125 & 0.0000 & 0 & 0.006 & 0.68 & 0\\
Llama 3.1 8B & 125 & 0.0000 & 0 & 0.001 & 0.01 & 0\\
Mistral 7B & 125 & 0.0080 & 0 & 0.001 & 0.01 & 0\\
Qwen2.5 7B & 125 & 0.0028 & 0 & 0.090 & 0.45 & 1\\
\bottomrule
\end{tabular}

\end{table}

\subsection{Tree-structured evaluation}
All prompts for one transcript and view share the transcript as a prefix. We encoded the shared prefix once, then evaluated the continuations (templates, formats and items) in a few forward passes. An attention mask restricted each continuation to its own ancestors in the prompt tree. Position indices continued from that continuation's prefix. For Transformers models this was a four-dimensional attention mask; for llama.cpp, one sequence per continuation sharing the cached prefix. Against one-prompt-at-a-time evaluation, the log-probabilities agreed to $10^{-4}$ in 32-bit arithmetic; in bfloat16 option probabilities differed by at most 0.03, and with llama.cpp log-probabilities agreed within 0.007. The checks covered one participant and 16 sampled continuations (the 20 most probable tokens of each). They diagnose numerical agreement without guaranteeing it for every prompt; Supplementary Table~\ref{tab:rescore} shows the reproducibility of whole scores across batch compositions.

\subsection{Generalizability analysis}
The data were the expected item sums $X_{pmtfv}$ for person $p$, model $m$, template $t$, format $f$ and view $v$, in a fully crossed design with one observation per cell. We computed mean squares for all 31 effects and solved their expectations for the variance components, working downwards from the highest-order interaction. Negative estimates were retained when solving for lower-order components, then set to zero for reporting\cite{Brennan2001}. The highest-order interaction is confounded with residual error. For a decision study that averages over $n'_m$ models, $n'_t$ templates and $n'_f$ formats (and, in discovery, over both views unless stated), the relative error variance $\sigma^2_\delta$ is the sum of the interactions involving the person, each divided by the numbers of conditions averaged over its other facets; the absolute error variance $\sigma^2_\Delta$ adds the effects not involving the person, divided in the same way. The coefficients are $E\rho^2 = \sigma^2_p/(\sigma^2_p + \sigma^2_\delta)$ and $\Phi = \sigma^2_p/(\sigma^2_p + \sigma^2_\Delta)$.

\subsection{Consistency indices}
For an item-sum configuration with answer distributions $p_{ik}$ for participant $i$ and item $k$, the expected total is $\mu_i = \sum_k E[X_{ik}]$ and, with items answered independently at temperature 1, the sampling variance of the total is $s^2_i = \sum_k \mathrm{Var}(X_{ik})$. The expected correlation between two independent sampled runs is $\mathrm{Var}_i(\mu_i)/[\mathrm{Var}_i(\mu_i) + \overline{s^2_i}]$. Cronbach's $\alpha$ was computed across the eight expected item scores. Agreement across templates is the mean Pearson correlation of the configuration's expected totals with those of the same model, format and view under each of the other four templates.

\subsection{Remedies}
For calibration, we randomly drew $n$ participants ($n = 20$ or 40) and fitted isotonic regression from each configuration's score to their PHQ-8 totals. We applied the fitted map to the remaining participants and evaluated the resulting decisions and implied prevalence. This was repeated for 200 draws. Prediction-powered estimates of prevalence used the configuration's decisions on all participants and the labels of the same $n$ participants to correct the bias of the model-based prevalence, with the standard variance estimate\cite{Angelopoulos2023}. The standard estimate targets a superpopulation. Our coverage assessment instead used the prevalence of the fixed cohort. In the exploratory analyses we therefore also computed a finite-cohort version, $\hat\theta = \bar f + \overline{(y-f)}_L$ with variance $(1-n/N)\,s^2_{y-f}/n$, and an estimate from the labels alone, $\bar y_L$ with variance $(1-n/N)\,s^2_y/n$, where $f$ is the configuration's decision, $y$ the label, $L$ the labelled participants and $N$ the cohort size.

\subsection{Decision analyses (exploratory)}
\paragraph{How many versus who.} For two raters with decisions $A$ and $B$ and positive rates $p_A$ and $p_B$, $P(A\neq B) = P(A{=}1,B{=}0) + P(A{=}0,B{=}1)$ and $p_A - p_B = P(A{=}1,B{=}0) - P(A{=}0,B{=}1)$, so $P(A\neq B) = |p_A-p_B| + 2\min\{P(A{=}1,B{=}0),P(A{=}0,B{=}1)\}$. The first term is the least disagreement that the two positive rates force; the second, the excess, is zero exactly when one rater's positives are a subset of the other's. We average both over all pairs of distinct raters in a set. The identity is elementary and assigns no causes: a pure difference of scale between two raters with identical rankings and equal means produces disagreement at a fixed threshold, so the residual after subtracting each rater's mean bias is not a measure of ranking disagreement. At equal capacity, every rater selects its $q$ highest-scoring participants; we used expected-value scores, whose boundaries had no ties at the capacities reported (ties would be broken by participant order and are counted in the code). Single-choice comparisons use every pair of configurations that are identical except in the named choice, at the threshold of 10 and, for expected-value pairs, at equal capacity.

\paragraph{Recalibration.} In each of 200 random splits (the same splits for every rule and rater set), 20, 40 or 80 participants formed a labelled sample (20 or 40 in confirmation) and the rest were held out. Three rater sets were recalibrated: raters with AUC $\geq$ 0.70 in the labelled sample; one fixed configuration per model (instrument wording, digits, participant's speech, expected value), which uses no selection; and, for comparison, raters with AUC $\geq$ 0.70 in the whole cohort, whose admission uses the held-out labels. The rules were: subtracting the mean difference between the score and the PHQ-8 total in the labelled sample; linear regression of the PHQ-8 total on the score, thresholded at 10; logistic regression of PHQ-8 $\geq$ 10 on the standardised score with a weak ridge penalty ($10^{-3}$, intercept included; Newton's method, convergence recorded), deciding at a probability of 0.5; the threshold, among the attainable ones (the distinct labelled scores), whose labelled positive count is closest to the number of labelled positives, compared exactly as integers, the higher threshold on equal distance; and the highest threshold keeping at least 80\% of the labelled positives, that is, the $\lceil 0.8\,n_+\rceil$-th largest labelled positive score. Ties among scores can make the prevalence target unattainable, and the achieved labelled rates are reported. Every rule sets a cut-off on the rater's own score. An increasing fitted relation preserves its ranking, a decreasing fit reverses it, and a constant fit flags everyone or no one. With 40 labels and raters admitted on the labelled sample, the linear fit was increasing in all but \RCLinNonInc{} of \RCLinFits{} rater--split combinations in discovery and in all \RCLinFitsC{} in confirmation. On the held-out participants we computed the chance that two recalibrated raters disagree, accuracy, the median sensitivity and specificity across raters, and the median absolute error of the share flagged, and averaged over splits; the central 95\% range across splits (2.5th to 97.5th percentiles) describes split-to-split variation within one cohort and is not a confidence interval (Supplementary Table~\ref{tab:recal}). Deferral set aside the held-out participants flagged by more than 20\% and fewer than 80\% of the recalibrated raters, retaining those at exactly 20\% or 80\%, and scored the majority decision on the rest.

\paragraph{Defensible core.} Besides the full grid, raters with AUC $\geq$ 0.70 and one fixed configuration per model, we report a clinically defensible core: the questionnaire's own instruction or the psychiatrist wording (neither changes the assessment policy), digit answers, expected values and symptom sums, in every view (44 raters in discovery, 22 in confirmation). It uses no outcome labels to choose raters (Supplementary Table~\ref{tab:restrict}).

\paragraph{Where disagreement falls.} For each participant, the chance that two raters decide differently, $2\pi_i(1-\pi_i)K/(K-1)$, was averaged within bands of self-reported PHQ-8 (0--4, 5--9, 10--14, 15--24), at the threshold of 10 and at equal capacity, and each band's share of the total was computed (Supplementary Table~\ref{tab:exchange}). This describes the criterion status of the participants on whom raters differ; it does not identify which decision is clinically right.

\paragraph{Deferral at matched coverage.} On the held-out participants of each split, with raters admitted on the 40 labels and recalibrated to the labelled prevalence, we compared on the same participants: a single recalibrated rater (averaged over raters); the majority of raters; the majority after deferring participants on whom fewer than a share $c$ of raters agreed ($c$ = 0.6, 0.7, 0.8, 0.9, 1.0; $c$ = 0.8 defers those flagged by strictly more than 20\% and fewer than 80\% of raters, retaining those at exactly 20\% or 80\%; unanimous consensus retained fewer than 3\% of participants and is not displayed); single raters evaluated on the same retained participants, which separates the effect of deferral from that of pooling; and each single rater deferring its own closest calls, the participants whose scores lie nearest its cut-off, to reach the same coverage. We report accuracy, sensitivity, specificity and the numbers of retained participants with and without a positive screen (Supplementary Table~\ref{tab:remedy}; Extended Data Fig.~9). Neither deferral rule has defined costs for review or for missed cases, so these are descriptive comparisons, not a clinical policy evaluation.

\paragraph{Uncertainty for headline estimands.} For three headline estimands (pairwise disagreement at 10 among raters with AUC $\geq$ 0.70, and equal-capacity disagreement among expected-value raters with AUC $\geq$ 0.70 and among one fixed configuration per model) we resampled participants 1,000 times with the rater sets held fixed; the percentile intervals are therefore conditional on the raters and describe sampling of participants, not of configurations.

\paragraph{Transfer.} For configurations with AUC $\geq$ 0.70 in discovery (participant view), three discovery calibrations were applied unchanged to the same configuration in confirmation: subtracting the discovery mean over-rating; the attainable discovery threshold closest to the discovery prevalence; and the discovery threshold keeping at least 80\% of discovery positives. The discovery set had manual transcripts and an operator-controlled interviewer, and the confirmation set automatic transcripts and an autonomous interviewer, so this is one related transfer, not a comparison of independent sites.

\begin{table}[H]
\small
\centering
\caption{Recalibration with 40 labelled participants, evaluated on the held-out participants (raters admitted by their AUC in the labelled sample; means over 200 splits). Disagreement, the chance that two recalibrated raters decide a held-out participant differently; error, median absolute error of the share flagged (2.5th--97.5th percentile across splits); target, the share of raters and splits whose labelled positive rate came within 5 points of the labelled prevalence (prevalence rule) or that kept at least 80\% of labelled positives (sensitivity rule).}
\label{tab:recal}

\setlength{\tabcolsep}{4pt}
\begin{tabular}{@{}llrrrrlr@{}}
\toprule
Set & Rule & Disagree (\%) & Acc. (\%) & Sens. (\%) & Spec. (\%) & Error (points) & Target (\%)\\
\midrule
Discovery & Threshold of 10 & 40 & 61 & 83 & 67 & 28 (21--32) & \\
 & Mean over-rating & 18 & 75 & 39 & 92 & 14 (7--25) & \\
 & Linear & 16 & 75 & 42 & 91 & 12 (5--27) & \\
 & Logistic & 17 & 75 & 45 & 89 & 11 (4--28) & \\
 & Labelled prevalence & 20 & 75 & 56 & 84 & 7 (3--19) & 88\\
 & $\geq$ 80\% of positives & 27 & 68 & 78 & 68 & 17 (7--33) & 100\\
Confirmation & Threshold of 10 & 40 & 59 & 91 & 53 & 33 (26--39) & \\
 & Mean over-rating & 21 & 74 & 49 & 92 & 14 (7--24) & \\
 & Linear & 20 & 75 & 54 & 89 & 11 (4--24) & \\
 & Logistic & 20 & 75 & 56 & 87 & 10 (4--22) & \\
 & Labelled prevalence & 21 & 75 & 62 & 84 & 8 (4--15) & 86\\
 & $\geq$ 80\% of positives & 27 & 69 & 79 & 70 & 15 (7--28) & 100\\
\bottomrule
\end{tabular}

\end{table}

\begin{table}[H]
\small
\centering
\caption{How the decision results change under restrictions of the grid. $K$, raters; at 10, the chance that two raters decide a participant differently at the threshold; forced, the share of that disagreement forced by different positive rates; equal no., the chance at equal capacity among expected-value raters of the set (all expected-value raters for the full grid); recal. and acc., held-out disagreement and accuracy after matching the labelled prevalence with 40 labels (no selection for the defensible core and the fixed configurations; whole-cohort AUC for the AUC $\geq$ 0.70 row).}
\label{tab:restrict}

\setlength{\tabcolsep}{4pt}
\begin{tabular}{@{}llrrrrrr@{}}
\toprule
Set & Raters & $K$ & At 10 (\%) & Forced (\%) & Equal no. (\%) & Recal. (\%) & Acc. (\%)\\
\midrule
Discovery & Full grid & 880 & 44 & 92 & 27 & -- & --\\
 & AUC $\geq$ 0.70 & 540 & 40 & 94 & 18 & 19 & 75\\
 & Defensible core & 44 & 31 & 95 & 21 & 21 & 74\\
 & One per model & 11 & 32 & 98 & 25 & 24 & 72\\
Confirmation & Full grid & 440 & 43 & 94 & 26 & -- & --\\
 & AUC $\geq$ 0.70 & 319 & 40 & 95 & 20 & 21 & 76\\
 & Defensible core & 22 & 30 & 97 & 21 & 20 & 76\\
 & One per model & 11 & 32 & 97 & 22 & 21 & 75\\
\bottomrule
\end{tabular}

\end{table}

\begin{table}[H]
\small
\centering
\caption{Where the disagreement falls, by self-reported PHQ-8, for expected-value raters with AUC $\geq$ 0.70: each participant's chance of being decided differently by two raters, averaged within the band (mean) and as the band's share of all such disagreement (share), at the threshold of 10 and at equal capacity.}
\label{tab:exchange}

\begin{tabular}{@{}llrrrrr@{}}
\toprule
 & & & \multicolumn{2}{l}{At 10} & \multicolumn{2}{l}{Equal numbers}\\
Set & PHQ-8 & $n$ & Mean (\%) & Share (\%) & Mean (\%) & Share (\%)\\
\midrule
Discovery & 0--4 & 86 & 44 & 48 & 9 & 24\\
 & 5--9 & 46 & 43 & 25 & 25 & 33\\
 & 10--14 & 30 & 40 & 15 & 27 & 24\\
 & 15--24 & 27 & 33 & 11 & 25 & 20\\
Confirmation & 0--4 & 36 & 47 & 48 & 13 & 28\\
 & 5--9 & 21 & 42 & 25 & 24 & 29\\
 & 10--14 & 13 & 38 & 14 & 30 & 23\\
 & 15--24 & 16 & 27 & 12 & 22 & 21\\
\bottomrule
\end{tabular}

\end{table}

\begin{table}[H]
\small
\centering
\caption{Deferral at matched coverage on held-out participants (raters admitted on 40 labels and recalibrated to the labelled prevalence; means over 200 splits). Coverage, participants decided rather than deferred; counts in parentheses, mean retained participants with and without a positive screen.}
\label{tab:remedy}

\begin{tabular}{@{}llrrrr@{}}
\toprule
Set & Rule & Coverage (\%) & Accuracy (\%) & Sens. (\%) & Spec. (\%)\\
\midrule
Discovery & Single rater, all & 100 & 75 & 55 & 83 \\
 & Majority, all & 100 & 78 & 58 & 88 \\
 & Majority, consensus $\geq$ 0.8 (24+, 82$-$) & 71 & 86 & 56 & 95 \\
 & \quad single rater, same participants & 71 & 83 & 54 & 91 \\
 & \quad single rater, own closest calls deferred & 71 & 82 & 48 & 92 \\
 & Majority, consensus $\geq$ 0.9 (13+, 70$-$) & 56 & 90 & 55 & 97 \\
 & \quad single rater, same participants & 56 & 88 & 54 & 94 \\
 & \quad single rater, own closest calls deferred & 56 & 85 & 45 & 94 \\
Confirmation & Single rater, all & 100 & 75 & 61 & 83 \\
 & Majority, all & 100 & 80 & 65 & 88 \\
 & Majority, consensus $\geq$ 0.8 (9+, 24$-$) & 70 & 89 & 70 & 95 \\
 & \quad single rater, same participants & 70 & 84 & 66 & 91 \\
 & \quad single rater, own closest calls deferred & 70 & 82 & 57 & 92 \\
 & Majority, consensus $\geq$ 0.9 (5+, 19$-$) & 53 & 92 & 65 & 98 \\
 & \quad single rater, same participants & 53 & 89 & 63 & 95 \\
 & \quad single rater, own closest calls deferred & 53 & 86 & 53 & 94 \\
\bottomrule
\end{tabular}

\end{table}

\begin{table}[H]
\small
\centering
\caption{Participants decided differently by configurations that differ in one choice only: median (interquartile range) over pairs, at the threshold of 10 and, for expected-value pairs, selecting as many participants as screen positive by self-report.}
\label{tab:single}

\begin{tabular}{@{}lllll@{}}
\toprule
 & \multicolumn{2}{l}{Discovery} & \multicolumn{2}{l}{Confirmation}\\
Choice changed & At 10 & Equal numbers & At 10 & Equal numbers\\
\midrule
Letter–answer mapping reversed & 20 (7--55) & 21 (12--43) & 26 (6--49) & 21 (12--36)\\
Digits vs letters & 13 (5--25) & 15 (7--24) & 12 (5--22) & 14 (9--22)\\
Psychiatrist persona & 7 (2--13) & 8 (6--15) & 6 (2--12) & 10 (7--16)\\
Unmentioned symptoms rated absent & 21 (8--35) & 14 (8--23) & 23 (9--36) & 14 (9--21)\\
Role-play as the participant & 8 (2--14) & 11 (7--17) & 8 (3--14) & 10 (7--19)\\
Use tone and other cues & 8 (2--17) & 11 (7--20) & 9 (2--14) & 14 (7--19)\\
Most likely option, not expected value & 4 (1--8) & -- & 3 (1--8) & --\\
One overall rating, not a symptom sum & 24 (10--49) & 15 (8--22) & 21 (10--35) & 16 (9--23)\\
Whole conversation, not participant's speech & 11 (3--19) & 17 (13--26) & -- & --\\
Next model size, same family & 28 (13--47) & 22 (15--33) & 30 (16--48) & 23 (16--30)\\
A different model & 35 (17--63) & 25 (17--33) & 35 (15--60) & 23 (16--33)\\
\bottomrule
\end{tabular}

\end{table}

\section{Supplementary Results}

\begin{table}[H]
\small
\caption{Results by model (discovery set; all 80 configurations of each model). Sens. and Spec., median sensitivity and specificity for PHQ-8 $\geq$ 10 over the model's configurations (descriptive). Bias, median over configurations of the mean difference between the configuration's score and the PHQ-8 total. Option mass, median over participants, templates and formats.}
\label{tab:bymodel}

\setlength{\tabcolsep}{5pt}
\begin{tabular}{@{}llrlrrrr@{}}
\toprule
Model & AUC, median (range) & AUC $\geq$ 0.70 (\%) & Positive, \% (range) & Sens. (\%) & Spec. (\%) & Bias & Option mass\\
\midrule
Qwen2.5 1.5B & 0.59 (0.41--0.77) & 8 & 50 (0--100) & 55 & 52 & +3.2 & 0.99\\
Qwen2.5 3B & 0.70 (0.49--0.82) & 49 & 84 (0--100) & 96 & 21 & +5.0 & 1.00\\
Qwen2.5 7B & 0.81 (0.64--0.87) & 96 & 67 (17--98) & 93 & 45 & +5.0 & 1.00\\
Qwen2.5 14B & 0.85 (0.78--0.88) & 100 & 33 (1--82) & 71 & 83 & +0.7 & 1.00\\
R1-Distill 7B & 0.67 (0.35--0.81) & 32 & 80 (1--100) & 89 & 23 & +5.3 & 0.99\\
Llama 3.2 1B & 0.61 (0.34--0.78) & 24 & 99 (0--100) & 100 & 2 & +7.3 & 0.89\\
Llama 3.2 3B & 0.74 (0.45--0.82) & 60 & 97 (58--100) & 100 & 4 & +7.2 & 1.00\\
Llama 3.1 8B & 0.84 (0.76--0.88) & 100 & 24 (0--68) & 54 & 89 & -0.4 & 0.99\\
Phi-3.5 mini & 0.63 (0.49--0.81) & 26 & 79 (3--99) & 84 & 23 & +4.6 & 1.00\\
Mistral 7B & 0.80 (0.55--0.87) & 92 & 85 (21--100) & 98 & 21 & +6.0 & 1.00\\
gpt-oss 20B & 0.81 (0.59--0.85) & 88 & 14 (0--46) & 31 & 95 & -0.8 & 0.91\\
\bottomrule
\end{tabular}

\end{table}

\begin{table}[H]
\small
\centering
\caption{Variance components of the expected item sum (person $\times$ model $\times$ template $\times$ format $\times$ view).}
\label{tab:gstudy}

\begin{tabular}{@{}lrrr@{}}
\toprule
Source & $\hat\sigma^2$ (raw) & $\hat\sigma^2$ & Share (\%)\\
\midrule
model & 6.943 & 6.943 & 30.0\\
model $\times$ format & 3.645 & 3.645 & 15.8\\
person & 2.437 & 2.437 & 10.5\\
person $\times$ model & 1.872 & 1.872 & 8.1\\
template & 1.779 & 1.779 & 7.7\\
format & 1.363 & 1.363 & 5.9\\
model $\times$ template & 0.971 & 0.971 & 4.2\\
model $\times$ template $\times$ format & 0.691 & 0.691 & 3.0\\
person $\times$ model $\times$ format $\times$ view & 0.588 & 0.588 & 2.5\\
person $\times$ model $\times$ format & 0.498 & 0.498 & 2.2\\
person $\times$ model $\times$ template $\times$ format $\times$ view & 0.429 & 0.429 & 1.9\\
person $\times$ format & 0.413 & 0.413 & 1.8\\
person $\times$ model $\times$ template & 0.244 & 0.244 & 1.1\\
model $\times$ format $\times$ view & 0.234 & 0.234 & 1.0\\
person $\times$ model $\times$ template $\times$ format & 0.215 & 0.215 & 0.9\\
person $\times$ model $\times$ template $\times$ view & 0.156 & 0.156 & 0.7\\
person $\times$ model $\times$ view & 0.155 & 0.155 & 0.7\\
person $\times$ view & 0.143 & 0.143 & 0.6\\
model $\times$ template $\times$ format $\times$ view & 0.138 & 0.138 & 0.6\\
person $\times$ template & 0.076 & 0.076 & 0.3\\
template $\times$ format & 0.043 & 0.043 & 0.2\\
person $\times$ template $\times$ format & 0.021 & 0.021 & 0.1\\
person $\times$ template $\times$ view & 0.014 & 0.014 & 0.1\\
model $\times$ view & 0.014 & 0.014 & 0.1\\
model $\times$ template $\times$ view & 0.014 & 0.014 & 0.1\\
template $\times$ view & 0.014 & 0.014 & 0.1\\
person $\times$ format $\times$ view & 0.006 & 0.006 & 0.0\\
person $\times$ template $\times$ format $\times$ view & 0.002 & 0.002 & 0.0\\
template $\times$ format $\times$ view & -0.000 & 0.000 & 0.0\\
format $\times$ view & -0.010 & 0.000 & 0.0\\
view & -0.009 & 0.000 & 0.0\\
\bottomrule
\end{tabular}

\end{table}

\begin{table}[H]
\small
\centering
\caption{Decision studies (discovery; both transcript views averaged). For one view, averaging over all models, templates and formats gives $E\rho^2$ = \ErhoAllOneView{} and $\Phi$ = \PhiAllOneView{}; a single configuration with one view gives $\Phi$ = \PhiOneOneView{}.}
\label{tab:dstudy}

\begin{tabular}{@{}rrrrr@{}}
\toprule
Models & Templates & Formats & $E\rho^2$ & $\Phi$\\
\midrule
1 & 1 & 1 & 0.37 & 0.11\\
3 & 1 & 1 & 0.58 & 0.21\\
11 & 1 & 1 & 0.73 & 0.32\\
1 & 1 & 3 & 0.45 & 0.14\\
3 & 1 & 3 & 0.67 & 0.27\\
11 & 1 & 3 & 0.82 & 0.40\\
1 & 5 & 1 & 0.42 & 0.13\\
3 & 5 & 1 & 0.62 & 0.26\\
11 & 5 & 1 & 0.76 & 0.41\\
1 & 5 & 3 & 0.49 & 0.17\\
3 & 5 & 3 & 0.71 & 0.34\\
11 & 5 & 3 & 0.85 & 0.55\\
\bottomrule
\end{tabular}

\end{table}

\begin{table}[H]
\small
\caption{Directional predictions specified before the analysis, and their outcomes.}
\label{tab:predictions}

\begin{tabularx}{\textwidth}{@{}lXlllll@{}}
\toprule
 & Prediction & Criterion & \multicolumn{2}{l}{Discovery} & \multicolumn{2}{l}{Confirmation}\\
\midrule
H1 & Configuration contributes more variance to the AUC than participant sampling & ratio $>$ 1 & 10.6 & supported & 5.6 & supported\\
H2 & Participants with discordant decisions among acceptable configurations & $\geq$ 25\% & 100\% & supported & 100\% & supported\\
H3 & Range of implied prevalence across configurations & $\geq$ 30 pp & 100 pp & supported & 100 pp & supported\\
H4 & Person share of item-sum variance & $<$ 50\% & 10.5\% & supported & 13.0\% & supported\\
H5 & Rank correlation of Cronbach's $\alpha$ with the AUC & $<$ 0.30 & 0.02 & supported & -0.06 & supported\\
H6 & Median-model AUC from interviewer speech only & $>$ 0.60 & 0.59 & not supported & -- & --\\
H7 & Prevalence range after calibration with 40 labels, against before & at most half & 58 vs 100 pp & not supported & 65 vs 100 pp & not supported\\
\bottomrule
\end{tabularx}

\end{table}

\section{Labels and transcripts}
All analyses use the registered definition of a positive screen, a PHQ-8 total of 10 or more. It differs from the binary flag distributed with the corpora for one discovery participant and five confirmation participants. Item scores were available for 142 discovery participants and complete for 141. In six discovery transcripts, empty turns had been converted to the literal token `nan' (1--12 tokens per transcript); excluding these six participants changed the median AUC across configurations from 0.753 to 0.757, and the frozen loader was not changed.

\section{Deviations from the pre-registered plan}
\par\medskip\noindent \textbf{Regenerated outputs.} Two secondary outputs of the first run, for Qwen2.5 1.5B (the interviewer's speech for one participant and the automatic transcript for another), contained no finite probabilities, and the median over models used for the interviewer-speech prediction silently omitted that model. Both outputs were regenerated with the frozen code, giving finite outputs with option mass 0.995 and 0.990; the original files are kept. Re-running the frozen analysis left the primary results unchanged and changed the interviewer-speech prediction from supported (median AUC 0.61) to not supported (\AucInterviewer{}).
\par\medskip\noindent \textbf{Direct-total termination.} The frozen code applied the end-of-number factor only to numbers whose tokens begin a longer candidate, not to every number as described (Supplementary Table~\ref{tab:rescore}). The code was not changed; in the re-scored sample (four models, 25 participants, participant view), the effect on expected totals was at most \TotDeltaMax{} points, with no change of decision; this measures the defect in that sample rather than bounding it for other models or views.
\par\medskip\noindent \textbf{Test attribution and degrees of freedom.} The plan names Steiger's test for comparing a configuration's correlations with the PHQ-8 and with PTSD severity; the code implements the Meng--Rosenthal--Rubin form\cite{Meng1992}. Steiger's pooled form\cite{Steiger1980} and a paired participant bootstrap give the same shares of configurations (\ShareSpecificSteiger\% and \ShareSpecificBoot\% in discovery; \ShareSpecificSteigerC\% and \ShareSpecificBootC\% in confirmation). The partial Spearman tests of the regularities used $n-2$ degrees of freedom; with $n-3$, every regularity still meets its criterion (PTSD regularity in confirmation, $P$ = \RulePtsdPC{}).
\par\medskip\noindent \textbf{Decision-study views.} In discovery the decision studies average over both transcript views; values for one view are given with Supplementary Table~\ref{tab:dstudy}.
\par\medskip\noindent During data collection the analysis code was changed once so that interim runs on an incomplete set of outputs recorded the interviewer-speech prediction as missing instead of stopping. The change was reverted before the registered analyses, which used the code as registered.
\par\medskip\noindent The local copies of three model weights (Qwen2.5 1.5B, 3B and 7B) were lost during data collection and obtained again from the same repositories. Re-scoring two participants with the new copy of Qwen2.5 1.5B reproduced the earlier outputs exactly.
\par\medskip\noindent Scoring runs were interrupted and resumed several times, and the order in which models were run changed. Answers were read from next-token probabilities without sampling, and every participant was scored once per model and view; across batch compositions, scores agree to the tolerance in Supplementary Table~\ref{tab:rescore}.
\par\medskip\noindent The two addenda (a label-free rule and nine regularities) were specified after the discovery analysis and before any confirmation output was compared with the criterion, while confirmation outputs were being generated. Analyses added after the confirmation analysis are listed in Methods and are exploratory.

\clearpage
\bibliographystyle{naturemag}
\bibliography{references}

@misc{Baumann2025,
  doi = {10.48550/ARXIV.2509.08825},
  author = {Baumann, Joachim and Röttger, Paul and Urman, Aleksandra and Wendsjö, Albert and Plaza-del-Arco, Flor Miriam and Gruber, Johannes B. and Hovy, Dirk},
  title = {Large Language Model Hacking: Quantifying the Hidden Risks of Using {LLMs} for Text Annotation},
  year = {2025},
  howpublished = {Preprint at \url{https://arxiv.org/abs/2509.08825}}
}

@article{Kebe2026, title={{LLAMADRS}: Evaluating Open-Source {LLMs} on Real Clinical Interviews---To Reason or Not to Reason?}, volume={14}, ISSN={2307-387X}, DOI={10.1162/tacl.a.741}, journal={Transactions of the Association for Computational Linguistics}, publisher={MIT Press}, author={Kebe, Gaoussou Youssouf and Girard, Jeffrey M. and Liebenthal, Einat and Baker, Justin and {De la Torre}, Fernando and Morency, Louis-Philippe}, year={2026}, pages={1589--1604},
  url = {https://aclanthology.org/2026.tacl-1.71/}
}

@inproceedings{Burdisso2024, title={{DAIC-WOZ}: On the Validity of Using the Therapist’s prompts in Automatic Depression Detection from Clinical Interviews}, DOI={10.18653/v1/2024.clinicalnlp-1.8}, booktitle={Proceedings of the 6th Clinical Natural Language Processing Workshop}, publisher={Association for Computational Linguistics}, author={Burdisso, Sergio and Reyes-Ramírez, Ernesto and Villatoro-tello, Esaú and Sánchez-Vega, Fernando and Lopez Monroy, Adrian and Motlicek, Petr}, year={2024}, pages={82–90},
  url = {https://doi.org/10.18653/v1/2024.clinicalnlp-1.8}
}

@article{Teferra2025, title={Leveraging large language models for automated depression screening}, volume={4}, ISSN={2767-3170}, DOI={10.1371/journal.pdig.0000943}, number={7}, journal={PLOS Digital Health}, publisher={Public Library of Science (PLoS)}, author={Teferra, Bazen Gashaw and Perivolaris, Argyrios and Hsiang, Wei-Ni and Sidharta, Christian Kevin and Rueda, Alice and Parkington, Karisa and Wu, Yuqi and Soni, Achint and Samavi, Reza and Jetly, Rakesh and Zhang, Yanbo and Cao, Bo and Rambhatla, Sirisha and Krishnan, Sri and Bhat, Venkat}, editor={Lange, Max}, year={2025}, pages={e0000943},
  url = {https://doi.org/10.1371/journal.pdig.0000943}
}

@inproceedings{Rosenman2024, title={{LLM} Questionnaire Completion for Automatic Psychiatric Assessment}, DOI={10.18653/v1/2024.findings-emnlp.23}, booktitle={Findings of the Association for Computational Linguistics: EMNLP 2024}, publisher={Association for Computational Linguistics}, author={Rosenman, Gony and Hendler, Talma and Wolf, Lior}, year={2024}, pages={403–415},
  url = {https://doi.org/10.18653/v1/2024.findings-emnlp.23}
}

@misc{Galatzer2023,
  doi = {10.48550/ARXIV.2308.01834},
  author = {Galatzer-Levy, Isaac R. and McDuff, Daniel and Natarajan, Vivek and Karthikesalingam, Alan and Malgaroli, Matteo},
  title = {The Capability of Large Language Models to Measure Psychiatric Functioning},
  year = {2023},
  howpublished = {Preprint at \url{https://arxiv.org/abs/2308.01834}}
}

@article{Zhu2026psychosis, title={Evaluating large language models for assessment of psychosis risk}, volume={9}, ISSN={2398-6352}, DOI={10.1038/s41746-026-02928-4}, number={1}, journal={npj Digital Medicine}, publisher={Springer Science and Business Media LLC}, author={Zhu, Taiyu and Tashevski, Alexander and Taquet, Maxime and Azis, Matilda and Jani, Tia and Broome, Matthew R. and Kabir, Thomas and Minichino, Amedeo and Murray, Graham K. and Nour, Matthew M. and Singh, Ilina and Fusar-Poli, Paolo and Nevado-Holgado, Alejo and McGuire, Philip and Oliver, Dominic}, year={2026},
  pages = {554},
  url = {https://doi.org/10.1038/s41746-026-02928-4}
}

@article{Chen2026bprs, title={Using Large Language Models to Measure Symptom Severity Scores in Patients At-Risk for Schizophrenia}, volume={52}, ISSN={1745-1701}, DOI={10.1093/schbul/sbag084}, number={5}, journal={Schizophrenia Bulletin}, publisher={Oxford University Press (OUP)}, author={Chen, Andrew X and Horga, Guillermo and Escola, Sean}, year={2026},
  pages = {sbag084},
  url = {https://doi.org/10.1093/schbul/sbag084}
}

@inproceedings{Sclar2024,
  author = {Sclar, Melanie and Choi, Yejin and Tsvetkov, Yulia and Suhr, Alane},
  title = {Quantifying Language Models' Sensitivity to Spurious Features in Prompt Design or: How {I} learned to start worrying about prompt formatting},
  booktitle = {The Twelfth International Conference on Learning Representations},
  year = {2024},
  url = {https://openreview.net/forum?id=RIu5lyNXjT}
}

@inproceedings{Zheng2024,
  author = {Zheng, Chujie and Zhou, Hao and Meng, Fandong and Zhou, Jie and Huang, Minlie},
  title = {Large Language Models Are Not Robust Multiple Choice Selectors},
  booktitle = {The Twelfth International Conference on Learning Representations},
  year = {2024},
  url = {https://openreview.net/forum?id=shr9PXz7T0}
}

@article{Mizrahi2024,
  author = {Mizrahi, Moran and Kaplan, Guy and Malkin, Dan and Dror, Rotem and Shahaf, Dafna and Stanovsky, Gabriel},
  title = {State of What Art? A Call for Multi-Prompt {LLM} Evaluation},
  journal = {Transactions of the Association for Computational Linguistics},
  volume = {12}, pages = {933--949}, year = {2024},
  doi = {10.1162/tacl_a_00681},
  url = {https://doi.org/10.1162/tacl_a_00681}
}

@misc{Atil2024,
  doi = {10.48550/ARXIV.2408.04667},
  author = {Atil, Berk and Aykent, Sarp and Chittams, Alexa and Fu, Lisheng and Passonneau, Rebecca J. and Radcliffe, Evan and Rajagopal, Guru Rajan and Sloan, Adam and Tudrej, Tomasz and Ture, Ferhan and Wu, Zhe and Xu, Lixinyu and Baldwin, Breck},
  title = {Non-Determinism of "Deterministic" {LLM} Settings},
  year = {2024},
  howpublished = {Preprint at \url{https://arxiv.org/abs/2408.04667}}
}

@misc{Norman2026,
  doi = {10.48550/ARXIV.2606.19544},
  author = {Norman, Justin D. and Rivera, Michael U. and Hughes, D. Alex},
  title = {Reliability without Validity: A Systematic, Large-Scale Evaluation of {LLM-as-a-Judge} Models Across Agreement, Consistency, and Bias},
  year = {2026},
  howpublished = {Preprint at \url{https://arxiv.org/abs/2606.19544}}
}

@misc{Batzdorfer2026,
  doi = {10.48550/ARXIV.2609.19866},
  author = {Batzdorfer, Veronika and Santagiustina, Carlo Romano Marcello Alessandro},
  title = {Reproducibility is not construct validity: {LLM} measurement of institutionally situated communication},
  year = {2026},
  howpublished = {Preprint at \url{https://arxiv.org/abs/2609.19866}}
}

@article{Obradovich2024, title={Opportunities and risks of large language models in psychiatry}, volume={2}, ISSN={2948-1570}, DOI={10.1038/s44277-024-00010-z}, number={1}, journal={NPP—Digital Psychiatry and Neuroscience}, publisher={Springer Science and Business Media LLC}, author={Obradovich, Nick and Khalsa, Sahib S. and Khan, Waqas U. and Suh, Jina and Perlis, Roy H. and Ajilore, Olusola and Paulus, Martin P.}, year={2024},
  pages = {8},
  url = {https://doi.org/10.1038/s44277-024-00010-z}
}

@article{Stade2024, title={Large language models could change the future of behavioral healthcare: a proposal for responsible development and evaluation}, volume={3}, ISSN={2731-4251}, DOI={10.1038/s44184-024-00056-z}, number={1}, journal={npj Mental Health Research}, publisher={Springer Science and Business Media LLC}, author={Stade, Elizabeth C. and Stirman, Shannon Wiltsey and Ungar, Lyle H. and Boland, Cody L. and Schwartz, H. Andrew and Yaden, David B. and Sedoc, João and DeRubeis, Robert J. and Willer, Robb and Eichstaedt, Johannes C.}, year={2024},
  pages = {12},
  url = {https://doi.org/10.1038/s44184-024-00056-z}
}

@article{Demszky2023, title={Using large language models in psychology}, ISSN={2731-0574}, DOI={10.1038/s44159-023-00241-5}, journal={Nature Reviews Psychology}, publisher={Springer Science and Business Media LLC}, author={Demszky, Dorottya and Yang, Diyi and Yeager, David S. and Bryan, Christopher J. and Clapper, Margarett and Chandhok, Susannah and Eichstaedt, Johannes C. and Hecht, Cameron and Jamieson, Jeremy and Johnson, Meghann and Jones, Michaela and Krettek-Cobb, Danielle and Lai, Leslie and JonesMitchell, Nirel and Ong, Desmond C. and Dweck, Carol S. and Gross, James J. and Pennebaker, James W.}, year={2023},
  url = {https://doi.org/10.1038/s44159-023-00241-5}
}

@article{Gallifant2025, title={The {TRIPOD-LLM} reporting guideline for studies using large language models}, volume={31}, ISSN={1546-170X}, DOI={10.1038/s41591-024-03425-5}, number={1}, journal={Nature Medicine}, publisher={Springer Science and Business Media LLC}, author={Gallifant, Jack and Afshar, Majid and Ameen, Saleem and Aphinyanaphongs, Yindalon and Chen, Shan and Cacciamani, Giovanni and Demner-Fushman, Dina and Dligach, Dmitriy and Daneshjou, Roxana and Fernandes, Chrystinne and Hansen, Lasse Hyldig and Landman, Adam and Lehmann, Lisa and McCoy, Liam G. and Miller, Timothy and Moreno, Amy and Munch, Nikolaj and Restrepo, David and Savova, Guergana and Umeton, Renato and Gichoya, Judy Wawira and Collins, Gary S. and Moons, Karel G. M. and Celi, Leo A. and Bitterman, Danielle S.}, year={2025}, pages={60–69},
  url = {https://doi.org/10.1038/s41591-024-03425-5}
}

@article{Angelopoulos2023, title={Prediction-powered inference}, volume={382}, ISSN={1095-9203}, DOI={10.1126/science.adi6000}, number={6671}, journal={Science}, publisher={American Association for the Advancement of Science (AAAS)}, author={Angelopoulos, Anastasios N. and Bates, Stephen and Fannjiang, Clara and Jordan, Michael I. and Zrnic, Tijana}, year={2023}, pages={669–674},
  url = {https://doi.org/10.1126/science.adi6000}
}

@article{Lin2025compass, title={{COMPASS}: Computational mapping of patient-therapist alliance strategies with language modeling}, volume={15}, ISSN={2158-3188}, DOI={10.1038/s41398-025-03379-3}, number={1}, journal={Translational Psychiatry}, publisher={Springer Science and Business Media LLC}, author={Lin, Baihan and Bouneffouf, Djallel and Landa, Yulia and Jespersen, Rachel and Corcoran, Cheryl and Cecchi, Guillermo}, year={2025},
  pages = {166},
  url = {https://doi.org/10.1038/s41398-025-03379-3}
}

@article{Tjuatja2024, title={Do {LLMs} Exhibit Human-like Response Biases? A Case Study in Survey Design}, volume={12}, ISSN={2307-387X}, DOI={10.1162/tacl_a_00685}, journal={Transactions of the Association for Computational Linguistics}, publisher={MIT Press}, author={Tjuatja, Lindia and Chen, Valerie and Wu, Tongshuang and Talwalkwar, Ameet and Neubig, Graham}, year={2024}, pages={1011–1026},
  url = {https://doi.org/10.1162/tacl_a_00685}
}

@article{LinZ2026, title={From Prompts to Constructs: A Dual-Validity Framework for Large Language Model Research in Psychology}, ISSN={1545-2085}, DOI={10.1146/annurev-psych-100925-034807}, journal={Annual Review of Psychology}, publisher={Annual Reviews}, author={Lin, Zhicheng}, year={2026},
  url = {https://doi.org/10.1146/annurev-psych-100925-034807}
}

@article{Botvinik2020, title={Variability in the analysis of a single neuroimaging dataset by many teams}, volume={582}, ISSN={1476-4687}, DOI={10.1038/s41586-020-2314-9}, number={7810}, journal={Nature}, publisher={Springer Science and Business Media LLC}, author={Botvinik-Nezer, Rotem and Holzmeister, Felix and Camerer, Colin F. and Dreber, Anna and Huber, Juergen and Johannesson, Magnus and Kirchler, Michael and Iwanir, Roni and Mumford, Jeanette A. and Adcock, R. Alison and Avesani, Paolo and Baczkowski, Blazej M. and Bajracharya, Aahana and Bakst, Leah and Ball, Sheryl and Barilari, Marco and Bault, Nadège and Beaton, Derek and Beitner, Julia and Benoit, Roland G. and Berkers, Ruud M. W. J. and Bhanji, Jamil P. and Biswal, Bharat B. and Bobadilla-Suarez, Sebastian and Bortolini, Tiago and Bottenhorn, Katherine L. and Bowring, Alexander and Braem, Senne and Brooks, Hayley R. and Brudner, Emily G. and Calderon, Cristian B. and Camilleri, Julia A. and Castrellon, Jaime J. and Cecchetti, Luca and Cieslik, Edna C. and Cole, Zachary J. and Collignon, Olivier and Cox, Robert W. and Cunningham, William A. and Czoschke, Stefan and Dadi, Kamalaker and Davis, Charles P. and Luca, Alberto De and Delgado, Mauricio R. and Demetriou, Lysia and Dennison, Jeffrey B. and Di, Xin and Dickie, Erin W. and Dobryakova, Ekaterina and Donnat, Claire L. and Dukart, Juergen and Duncan, Niall W. and Durnez, Joke and Eed, Amr and Eickhoff, Simon B. and Erhart, Andrew and Fontanesi, Laura and Fricke, G. Matthew and Fu, Shiguang and Galván, Adriana and Gau, Remi and Genon, Sarah and Glatard, Tristan and Glerean, Enrico and Goeman, Jelle J. and Golowin, Sergej A. E. and González-García, Carlos and Gorgolewski, Krzysztof J. and Grady, Cheryl L. and Green, Mikella A. and Guassi Moreira, João F. and Guest, Olivia and Hakimi, Shabnam and Hamilton, J. Paul and Hancock, Roeland and Handjaras, Giacomo and Harry, Bronson B. and Hawco, Colin and Herholz, Peer and Herman, Gabrielle and Heunis, Stephan and Hoffstaedter, Felix and Hogeveen, Jeremy and Holmes, Susan and Hu, Chuan-Peng and Huettel, Scott A. and Hughes, Matthew E. and Iacovella, Vittorio and Iordan, Alexandru D. and Isager, Peder M. and Isik, Ayse I. and Jahn, Andrew and Johnson, Matthew R. and Johnstone, Tom and Joseph, Michael J. E. and Juliano, Anthony C. and Kable, Joseph W. and Kassinopoulos, Michalis and Koba, Cemal and Kong, Xiang-Zhen and Koscik, Timothy R. and Kucukboyaci, Nuri Erkut and Kuhl, Brice A. and Kupek, Sebastian and Laird, Angela R. and Lamm, Claus and Langner, Robert and Lauharatanahirun, Nina and Lee, Hongmi and Lee, Sangil and Leemans, Alexander and Leo, Andrea and Lesage, Elise and Li, Flora and Li, Monica Y. C. and Lim, Phui Cheng and Lintz, Evan N. and Liphardt, Schuyler W. and Losecaat Vermeer, Annabel B. and Love, Bradley C. and Mack, Michael L. and Malpica, Norberto and Marins, Theo and Maumet, Camille and McDonald, Kelsey and McGuire, Joseph T. and Melero, Helena and Méndez Leal, Adriana S. and Meyer, Benjamin and Meyer, Kristin N. and Mihai, Glad and Mitsis, Georgios D. and Moll, Jorge and Nielson, Dylan M. and Nilsonne, Gustav and Notter, Michael P. and Olivetti, Emanuele and Onicas, Adrian I. and Papale, Paolo and Patil, Kaustubh R. and Peelle, Jonathan E. and Pérez, Alexandre and Pischedda, Doris and Poline, Jean-Baptiste and Prystauka, Yanina and Ray, Shruti and Reuter-Lorenz, Patricia A. and Reynolds, Richard C. and Ricciardi, Emiliano and Rieck, Jenny R. and Rodriguez-Thompson, Anais M. and Romyn, Anthony and Salo, Taylor and Samanez-Larkin, Gregory R. and Sanz-Morales, Emilio and Schlichting, Margaret L. and Schultz, Douglas H. and Shen, Qiang and Sheridan, Margaret A. and Silvers, Jennifer A. and Skagerlund, Kenny and Smith, Alec and Smith, David V. and Sokol-Hessner, Peter and Steinkamp, Simon R. and Tashjian, Sarah M. and Thirion, Bertrand and Thorp, John N. and Tinghög, Gustav and Tisdall, Loreen and Tompson, Steven H. and Toro-Serey, Claudio and Torre Tresols, Juan Jesus and Tozzi, Leonardo and Truong, Vuong and Turella, Luca and van ’t Veer, Anna E. and Verguts, Tom and Vettel, Jean M. and Vijayarajah, Sagana and Vo, Khoi and Wall, Matthew B. and Weeda, Wouter D. and Weis, Susanne and White, David J. and Wisniewski, David and Xifra-Porxas, Alba and Yearling, Emily A. and Yoon, Sangsuk and Yuan, Rui and Yuen, Kenneth S. L. and Zhang, Lei and Zhang, Xu and Zosky, Joshua E. and Nichols, Thomas E. and Poldrack, Russell A. and Schonberg, Tom}, year={2020}, pages={84–88},
  url = {https://doi.org/10.1038/s41586-020-2314-9}
}

@article{Silberzahn2018, title={Many Analysts, One Data Set: Making Transparent How Variations in Analytic Choices Affect Results}, volume={1}, ISSN={2515-2467}, DOI={10.1177/2515245917747646}, number={3}, journal={Advances in Methods and Practices in Psychological Science}, publisher={SAGE Publications}, author={Silberzahn, R. and Uhlmann, E. L. and Martin, D. P. and Anselmi, P. and Aust, F. and Awtrey, E. and Bahník, Š. and Bai, F. and Bannard, C. and Bonnier, E. and Carlsson, R. and Cheung, F. and Christensen, G. and Clay, R. and Craig, M. A. and Dalla Rosa, A. and Dam, L. and Evans, M. H. and Flores Cervantes, I. and Fong, N. and Gamez-Djokic, M. and Glenz, A. and Gordon-McKeon, S. and Heaton, T. J. and Hederos, K. and Heene, M. and Hofelich Mohr, A. J. and Högden, F. and Hui, K. and Johannesson, M. and Kalodimos, J. and Kaszubowski, E. and Kennedy, D. M. and Lei, R. and Lindsay, T. A. and Liverani, S. and Madan, C. R. and Molden, D. and Molleman, E. and Morey, R. D. and Mulder, L. B. and Nijstad, B. R. and Pope, N. G. and Pope, B. and Prenoveau, J. M. and Rink, F. and Robusto, E. and Roderique, H. and Sandberg, A. and Schlüter, E. and Schönbrodt, F. D. and Sherman, M. F. and Sommer, S. A. and Sotak, K. and Spain, S. and Spörlein, C. and Stafford, T. and Stefanutti, L. and Tauber, S. and Ullrich, J. and Vianello, M. and Wagenmakers, E.-J. and Witkowiak, M. and Yoon, S. and Nosek, B. A.}, year={2018}, pages={337–356},
  url = {https://doi.org/10.1177/2515245917747646}
}

@article{Steegen2016, title={Increasing Transparency Through a Multiverse Analysis}, volume={11}, ISSN={1745-6924}, DOI={10.1177/1745691616658637}, number={5}, journal={Perspectives on Psychological Science}, publisher={SAGE Publications}, author={Steegen, Sara and Tuerlinckx, Francis and Gelman, Andrew and Vanpaemel, Wolf}, year={2016}, pages={702–712},
  url = {https://doi.org/10.1177/1745691616658637}
}

@article{Simonsohn2020, title={Specification curve analysis}, volume={4}, ISSN={2397-3374}, DOI={10.1038/s41562-020-0912-z}, number={11}, journal={Nature Human Behaviour}, publisher={Springer Science and Business Media LLC}, author={Simonsohn, Uri and Simmons, Joseph P. and Nelson, Leif D.}, year={2020}, pages={1208–1214},
  url = {https://doi.org/10.1038/s41562-020-0912-z}
}

@article{Gelman2014, title={Beyond Power Calculations: Assessing {Type} S (Sign) and {Type} M (Magnitude) Errors}, volume={9}, ISSN={1745-6924}, DOI={10.1177/1745691614551642}, number={6}, journal={Perspectives on Psychological Science}, publisher={SAGE Publications}, author={Gelman, Andrew and Carlin, John}, year={2014}, pages={641–651},
  url = {https://doi.org/10.1177/1745691614551642}
}

@book{Brennan2001, title={Generalizability Theory}, ISBN={9781475734560}, DOI={10.1007/978-1-4757-3456-0}, publisher={Springer New York}, author={Brennan, Robert L.}, year={2001},
  url = {https://doi.org/10.1007/978-1-4757-3456-0}
}

@article{Kroenke2009, title={The {PHQ-8} as a measure of current depression in the general population}, volume={114}, ISSN={0165-0327}, DOI={10.1016/j.jad.2008.06.026}, number={1-3}, journal={Journal of Affective Disorders}, publisher={Elsevier BV}, author={Kroenke, Kurt and Strine, Tara W. and Spitzer, Robert L. and Williams, Janet B.W. and Berry, Joyce T. and Mokdad, Ali H.}, year={2009}, pages={163–173},
  url = {https://doi.org/10.1016/j.jad.2008.06.026}
}

@inproceedings{Ringeval2019, series={MM ’19}, title={{AVEC} 2019 Workshop and Challenge: State-of-Mind, Detecting Depression with {AI}, and Cross-Cultural Affect Recognition}, DOI={10.1145/3347320.3357688}, booktitle={Proceedings of the 9th International on Audio/Visual Emotion Challenge and Workshop}, publisher={ACM}, author={Ringeval, Fabien and Schuller, Björn and Valstar, Michel and Cummins, Nicholas and Cowie, Roddy and Tavabi, Leili and Schmitt, Maximilian and Alisamir, Sina and Amiriparian, Shahin and Messner, Eva-Maria and Song, Siyang and Liu, Shuo and Zhao, Ziping and Mallol-Ragolta, Adria and Ren, Zhao and Soleymani, Mohammad and Pantic, Maja}, year={2019}, pages={3–12}, collection={MM ’19},
  url = {https://doi.org/10.1145/3347320.3357688}
}

@inproceedings{Valstar2016, series={MM ’16}, title={{AVEC} 2016: Depression, Mood, and Emotion Recognition Workshop and Challenge}, DOI={10.1145/2988257.2988258}, booktitle={Proceedings of the 6th International Workshop on Audio/Visual Emotion Challenge}, publisher={ACM}, author={Valstar, Michel and Gratch, Jonathan and Schuller, Björn and Ringeval, Fabien and Lalanne, Denis and Torres Torres, Mercedes and Scherer, Stefan and Stratou, Giota and Cowie, Roddy and Pantic, Maja}, year={2016}, pages={3–10}, collection={MM ’16},
  url = {https://doi.org/10.1145/2988257.2988258}
}

@article{Shrout1979, title={Intraclass correlations: Uses in assessing rater reliability.}, volume={86}, ISSN={0033-2909}, DOI={10.1037/0033-2909.86.2.420}, number={2}, journal={Psychological Bulletin}, publisher={American Psychological Association (APA)}, author={Shrout, Patrick E. and Fleiss, Joseph L.}, year={1979}, pages={420–428},
  url = {https://doi.org/10.1037/0033-2909.86.2.420}
}

@article{Steiger1980, title={Tests for comparing elements of a correlation matrix.}, volume={87}, ISSN={0033-2909}, DOI={10.1037/0033-2909.87.2.245}, number={2}, journal={Psychological Bulletin}, publisher={American Psychological Association (APA)}, author={Steiger, James H.}, year={1980}, pages={245–251},
  url = {https://doi.org/10.1037/0033-2909.87.2.245}
}

@article{Meng1992, title={Comparing correlated correlation coefficients.}, volume={111}, ISSN={0033-2909}, DOI={10.1037/0033-2909.111.1.172}, number={1}, journal={Psychological Bulletin}, publisher={American Psychological Association (APA)}, author={Meng, Xiao-li and Rosenthal, Robert and Rubin, Donald B.}, year={1992}, pages={172–175},
  url = {https://doi.org/10.1037/0033-2909.111.1.172}
}

@inproceedings{Du2025,
  author = {Du, Ally Yalei and Ngo, Dung Daniel and Wu, Zhiwei Steven},
  title = {Reconciling Model Multiplicity for Downstream Decision Making},
  booktitle = {The Thirteenth International Conference on Learning Representations},
  year = {2025},
  url = {https://openreview.net/forum?id=uy4EavBEwl}
}

@misc{Lockfisch2025,
  doi = {10.48550/ARXIV.2507.19408},
  author = {Lockfisch, Sarah and Schwethelm, Kristian and Menten, Martin and Braren, Rickmer and Rueckert, Daniel and Ziller, Alexander and Kaissis, Georgios},
  title = {On Arbitrary Predictions from Equally Valid Models},
  year = {2025},
  howpublished = {Preprint at \url{https://arxiv.org/abs/2507.19408}}
}

@article{MacCoun2015, title={Blind analysis: Hide results to seek the truth}, volume={526}, ISSN={1476-4687}, DOI={10.1038/526187a}, number={7572}, journal={Nature}, publisher={Springer Science and Business Media LLC}, author={MacCoun, Robert and Perlmutter, Saul}, year={2015}, pages={187–189},
  url = {https://doi.org/10.1038/526187a}
}

@article{Nosek2018, title={The preregistration revolution}, volume={115}, ISSN={1091-6490}, DOI={10.1073/pnas.1708274114}, number={11}, journal={Proceedings of the National Academy of Sciences}, publisher={National Academy of Sciences}, author={Nosek, Brian A. and Ebersole, Charles R. and DeHaven, Alexander C. and Mellor, David T.}, year={2018}, pages={2600–2606},
  url = {https://doi.org/10.1073/pnas.1708274114}
}

@article{Levis2019, title={Accuracy of Patient Health Questionnaire-9 ({PHQ-9}) for screening to detect major depression: individual participant data meta-analysis}, ISSN={1756-1833}, DOI={10.1136/bmj.l1476}, journal={BMJ}, publisher={BMJ}, author={Levis, Brooke and Benedetti, Andrea and Thombs, Brett D}, year={2019}, pages={l1476},
  url = {https://doi.org/10.1136/bmj.l1476}
}

@article{Hanley1982, title={The meaning and use of the area under a receiver operating characteristic ({ROC}) curve.}, volume={143}, ISSN={1527-1315}, DOI={10.1148/radiology.143.1.7063747}, number={1}, journal={Radiology}, publisher={Radiological Society of North America (RSNA)}, author={Hanley, J A and McNeil, B J}, year={1982}, pages={29–36},
  url = {https://doi.org/10.1148/radiology.143.1.7063747}
}

@article{Spitzer1974, title={A Re-analysis of the Reliability of Psychiatric Diagnosis}, volume={125}, ISSN={1472-1465}, DOI={10.1192/bjp.125.4.341}, number={587}, journal={British Journal of Psychiatry}, publisher={Royal College of Psychiatrists}, author={Spitzer, Robert L. and Fleiss, Joseph L.}, year={1974}, pages={341–347},
  url = {https://doi.org/10.1192/bjp.125.4.341}
}

@article{Kendell1971, title={Diagnostic Criteria of American and British Psychiatrists}, volume={25}, ISSN={0003-990X}, DOI={10.1001/archpsyc.1971.01750140027006}, number={2}, journal={Archives of General Psychiatry}, publisher={American Medical Association (AMA)}, author={Kendell, R. E.}, year={1971}, pages={123},
  url = {https://doi.org/10.1001/archpsyc.1971.01750140027006}
}

@article{Feighner1972, title={Diagnostic Criteria for Use in Psychiatric Research}, volume={26}, ISSN={0003-990X}, DOI={10.1001/archpsyc.1972.01750190059011}, number={1}, journal={Archives of General Psychiatry}, publisher={American Medical Association (AMA)}, author={Feighner, John P.}, year={1972}, pages={57},
  url = {https://doi.org/10.1001/archpsyc.1972.01750190059011}
}

@article{Spitzer1978, title={Research Diagnostic Criteria: Rationale and Reliability}, volume={35}, ISSN={0003-990X}, DOI={10.1001/archpsyc.1978.01770300115013}, number={6}, journal={Archives of General Psychiatry}, publisher={American Medical Association (AMA)}, author={Spitzer, Robert L.}, year={1978}, pages={773},
  url = {https://doi.org/10.1001/archpsyc.1978.01770300115013}
}

@misc{Lorenzoni2025,
  doi = {10.48550/ARXIV.2501.00199},
  author = {Lorenzoni, Giuliano and Velmovitsky, Pedro Elkind and Alencar, Paulo and Cowan, Donald},
  title = {GPT-4 on Clinic Depression Assessment: An LLM-Based Pilot Study},
  year = {2025},
  howpublished = {Preprint at \url{https://arxiv.org/abs/2501.00199}}
}

@misc{Alkaeed2026,
  doi = {10.48550/ARXIV.2605.30646},
  author = {Alkaeed, Mahdi and Qayyum, Adnan and Kashreef, Nabeel Abo and Bilal, Muhammad and Qadir, Junaid},
  title = {Same Patient, Different Words, Different Diagnosis? Evaluating Semantic Stability in Clinical {LLMs}},
  year = {2026},
  howpublished = {Preprint at \url{https://arxiv.org/abs/2605.30646}}
}

@misc{XuHu2026,
  doi = {10.48550/ARXIV.2603.06910},
  author = {Xu, Jiayi and Hu, Xiyang},
  title = {Language Shapes Mental Health Evaluations in Large Language Models},
  year = {2026},
  howpublished = {Preprint at \url{https://arxiv.org/abs/2603.06910}}
}

@inproceedings{Marx2020,
  author = {Marx, Charles T. and Calmon, Flavio du Pin and Ustun, Berk},
  title = {Predictive Multiplicity in Classification},
  booktitle = {Proceedings of the 37th International Conference on Machine Learning},
  year = {2020},
  url = {https://proceedings.mlr.press/v119/marx20a.html},
  volume = {119},
  pages = {6765--6774},
  series = {Proceedings of Machine Learning Research}
}

@inproceedings{Black2022, series={FAccT ’22}, title={Model Multiplicity: Opportunities, Concerns, and Solutions}, DOI={10.1145/3531146.3533149}, booktitle={2022 ACM Conference on Fairness Accountability and Transparency}, publisher={ACM}, author={Black, Emily and Raghavan, Manish and Barocas, Solon}, year={2022}, pages={850–863}, collection={FAccT ’22},
  url = {https://doi.org/10.1145/3531146.3533149}
}

@article{VanCalster2019, title={Calibration: the Achilles heel of predictive analytics}, volume={17}, ISSN={1741-7015}, DOI={10.1186/s12916-019-1466-7}, number={1}, journal={BMC Medicine}, publisher={Springer Science and Business Media LLC}, author={Van Calster, Ben and McLernon, David J. and van Smeden, Maarten and Wynants, Laure and Steyerberg, Ewout W. and {On behalf of Topic Group `Evaluating diagnostic tests and prediction models' of the STRATOS initiative}}, year={2019},
  pages = {230},
  url = {https://doi.org/10.1186/s12916-019-1466-7}
}

@article{Tierney2024, title={Ambient Artificial Intelligence Scribes to Alleviate the Burden of Clinical Documentation}, volume={5}, ISSN={2642-0007}, DOI={10.1056/cat.23.0404}, number={3}, journal={NEJM Catalyst}, publisher={Massachusetts Medical Society}, author={Tierney, Aaron A. and Gayre, Gregg and Hoberman, Brian and Mattern, Britt and Ballesca, Manuel and Kipnis, Patricia and Liu, Vincent and Lee, Kristine}, year={2024},
  url = {https://doi.org/10.1056/cat.23.0404}
}

@article{Pugh2024, title={Assessing dimensions of thought disorder with large language models: The tradeoff of accuracy and consistency}, volume={341}, ISSN={0165-1781}, DOI={10.1016/j.psychres.2024.116119}, journal={Psychiatry Research}, publisher={Elsevier BV}, author={Pugh, Samuel L. and Chandler, Chelsea and Cohen, Alex S. and Diaz-Asper, Catherine and Elvevåg, Brita and Foltz, Peter W.}, year={2024}, pages={116119},
  url = {https://doi.org/10.1016/j.psychres.2024.116119}
}

@misc{Qwen2025,
  doi = {10.48550/ARXIV.2412.15115},
  author = {{Qwen} and Yang, An and Yang, Baosong and Zhang, Beichen and Hui, Binyuan and Zheng, Bo and Yu, Bowen and Li, Chengyuan and Liu, Dayiheng and Huang, Fei and Wei, Haoran and Lin, Huan and Yang, Jian and Tu, Jianhong and Zhang, Jianwei and Yang, Jianxin and Yang, Jiaxi and Zhou, Jingren and Lin, Junyang and Dang, Kai and Lu, Keming and Bao, Keqin and Yang, Kexin and Yu, Le and Li, Mei and Xue, Mingfeng and Zhang, Pei and Zhu, Qin and Men, Rui and Lin, Runji and Li, Tianhao and Tang, Tianyi and Xia, Tingyu and Ren, Xingzhang and Ren, Xuancheng and Fan, Yang and Su, Yang and Zhang, Yichang and Wan, Yu and Liu, Yuqiong and Cui, Zeyu and Zhang, Zhenru and Qiu, Zihan},
  title = {{Qwen2.5} Technical Report},
  year = {2024},
  howpublished = {Preprint at \url{https://arxiv.org/abs/2412.15115}}
}

@misc{Llama3,
  doi = {10.48550/ARXIV.2407.21783},
  author = {Grattafiori, Aaron and Dubey, Abhimanyu and Jauhri, Abhinav and Pandey, Abhinav and Kadian, Abhishek and Al-Dahle, Ahmad and Letman, Aiesha and Mathur, Akhil and Schelten, Alan and Vaughan, Alex and Yang, Amy and Fan, Angela and Goyal, Anirudh and Hartshorn, Anthony and Yang, Aobo and Mitra, Archi and Sravankumar, Archie and Korenev, Artem and Hinsvark, Arthur and Rao, Arun and Zhang, Aston and Rodriguez, Aurelien and Gregerson, Austen and Spataru, Ava and Roziere, Baptiste and Biron, Bethany and Tang, Binh and Chern, Bobbie and Caucheteux, Charlotte and Nayak, Chaya and Bi, Chloe and Marra, Chris and McConnell, Chris and Keller, Christian and Touret, Christophe and Wu, Chunyang and Wong, Corinne and Ferrer, Cristian Canton and Nikolaidis, Cyrus and Allonsius, Damien and Song, Daniel and Pintz, Danielle and Livshits, Danny and Wyatt, Danny and Esiobu, David and Choudhary, Dhruv and Mahajan, Dhruv and Garcia-Olano, Diego and Perino, Diego and Hupkes, Dieuwke and Lakomkin, Egor and AlBadawy, Ehab and Lobanova, Elina and Dinan, Emily and Smith, Eric Michael and Radenovic, Filip and Guzmán, Francisco and Zhang, Frank and Synnaeve, Gabriel and Lee, Gabrielle and Anderson, Georgia Lewis and Thattai, Govind and Nail, Graeme and Mialon, Gregoire and Pang, Guan and Cucurell, Guillem and Nguyen, Hailey and Korevaar, Hannah and Xu, Hu and Touvron, Hugo and Zarov, Iliyan and Ibarra, Imanol Arrieta and Kloumann, Isabel and Misra, Ishan and Evtimov, Ivan and Zhang, Jack and Copet, Jade and Lee, Jaewon and Geffert, Jan and Vranes, Jana and Park, Jason and Mahadeokar, Jay and Shah, Jeet and van der Linde, Jelmer and Billock, Jennifer and Hong, Jenny and Lee, Jenya and Fu, Jeremy and Chi, Jianfeng and Huang, Jianyu and Liu, Jiawen and Wang, Jie and Yu, Jiecao and Bitton, Joanna and Spisak, Joe and Park, Jongsoo and Rocca, Joseph and Johnstun, Joshua and Saxe, Joshua and Jia, Junteng and Alwala, Kalyan Vasuden and Prasad, Karthik and Upasani, Kartikeya and Plawiak, Kate and Li, Ke and Heafield, Kenneth and Stone, Kevin and El-Arini, Khalid and Iyer, Krithika and Malik, Kshitiz and Chiu, Kuenley and Bhalla, Kunal and Lakhotia, Kushal and Rantala-Yeary, Lauren and van der Maaten, Laurens and Chen, Lawrence and Tan, Liang and Jenkins, Liz and Martin, Louis and Madaan, Lovish and Malo, Lubo and Blecher, Lukas and Landzaat, Lukas and de Oliveira, Luke and Muzzi, Madeline and Pasupuleti, Mahesh and Singh, Mannat and Paluri, Manohar and Kardas, Marcin and Tsimpoukelli, Maria and Oldham, Mathew and Rita, Mathieu and Pavlova, Maya and Kambadur, Melanie and Lewis, Mike and Si, Min and Singh, Mitesh Kumar and Hassan, Mona and Goyal, Naman and Torabi, Narjes and Bashlykov, Nikolay and Bogoychev, Nikolay and Chatterji, Niladri and Zhang, Ning and Duchenne, Olivier and Çelebi, Onur and Alrassy, Patrick and Zhang, Pengchuan and Li, Pengwei and Vasic, Petar and Weng, Peter and Bhargava, Prajjwal and Dubal, Pratik and Krishnan, Praveen and Koura, Punit Singh and Xu, Puxin and He, Qing and Dong, Qingxiao and Srinivasan, Ragavan and Ganapathy, Raj and Calderer, Ramon and Cabral, Ricardo Silveira and Stojnic, Robert and Raileanu, Roberta and Maheswari, Rohan and Girdhar, Rohit and Patel, Rohit and Sauvestre, Romain and Polidoro, Ronnie and Sumbaly, Roshan and Taylor, Ross and Silva, Ruan and Hou, Rui and Wang, Rui and Hosseini, Saghar and Chennabasappa, Sahana and Singh, Sanjay and Bell, Sean and Kim, Seohyun Sonia and Edunov, Sergey and Nie, Shaoliang and Narang, Sharan and Raparthy, Sharath and Shen, Sheng and Wan, Shengye and Bhosale, Shruti and Zhang, Shun and Vandenhende, Simon and Batra, Soumya and Whitman, Spencer and Sootla, Sten and Collot, Stephane and Gururangan, Suchin and Borodinsky, Sydney and Herman, Tamar and Fowler, Tara and Sheasha, Tarek and Georgiou, Thomas and Scialom, Thomas and Speckbacher, Tobias and Mihaylov, Todor and Xiao, Tong and Karn, Ujjwal and Goswami, Vedanuj and Gupta, Vibhor and Ramanathan, Vignesh and Kerkez, Viktor and Gonguet, Vincent and Do, Virginie and Vogeti, Vish and Albiero, Vítor and Petrovic, Vladan and Chu, Weiwei and Xiong, Wenhan and Fu, Wenyin and Meers, Whitney and Martinet, Xavier and Wang, Xiaodong and Wang, Xiaofang and Tan, Xiaoqing Ellen and Xia, Xide and Xie, Xinfeng and Jia, Xuchao and Wang, Xuewei and Goldschlag, Yaelle and Gaur, Yashesh and Babaei, Yasmine and Wen, Yi and Song, Yiwen and Zhang, Yuchen and Li, Yue and Mao, Yuning and Coudert, Zacharie Delpierre and Yan, Zheng and Chen, Zhengxing and Papakipos, Zoe and Singh, Aaditya and Srivastava, Aayushi and Jain, Abha and Kelsey, Adam and Shajnfeld, Adam and Gangidi, Adithya and Victoria, Adolfo and Goldstand, Ahuva and Menon, Ajay and Sharma, Ajay and Boesenberg, Alex and Baevski, Alexei and Feinstein, Allie and Kallet, Amanda and Sangani, Amit and Teo, Amos and Yunus, Anam and Lupu, Andrei and Alvarado, Andres and Caples, Andrew and Gu, Andrew and Ho, Andrew and Poulton, Andrew and Ryan, Andrew and Ramchandani, Ankit and Dong, Annie and Franco, Annie and Goyal, Anuj and Saraf, Aparajita and Chowdhury, Arkabandhu and Gabriel, Ashley and Bharambe, Ashwin and Eisenman, Assaf and Yazdan, Azadeh and James, Beau and Maurer, Ben and Leonhardi, Benjamin and Huang, Bernie and Loyd, Beth and De Paola, Beto and Paranjape, Bhargavi and Liu, Bing and Wu, Bo and Ni, Boyu and Hancock, Braden and Wasti, Bram and Spence, Brandon and Stojkovic, Brani and Gamido, Brian and Montalvo, Britt and Parker, Carl and Burton, Carly and Mejia, Catalina and Liu, Ce and Wang, Changhan and Kim, Changkyu and Zhou, Chao and Hu, Chester and Chu, Ching-Hsiang and Cai, Chris and Tindal, Chris and Feichtenhofer, Christoph and Gao, Cynthia and Civin, Damon and Beaty, Dana and Kreymer, Daniel and Li, Daniel and Adkins, David and Xu, David and Testuggine, Davide and David, Delia and Parikh, Devi and Liskovich, Diana and Foss, Didem and Wang, Dingkang and Le, Duc and Holland, Dustin and Dowling, Edward and Jamil, Eissa and Montgomery, Elaine and Presani, Eleonora and Hahn, Emily and Wood, Emily and Le, Eric-Tuan and Brinkman, Erik and Arcaute, Esteban and Dunbar, Evan and Smothers, Evan and Sun, Fei and Kreuk, Felix and Tian, Feng and Kokkinos, Filippos and Ozgenel, Firat and Caggioni, Francesco and Kanayet, Frank and Seide, Frank and Florez, Gabriela Medina and Schwarz, Gabriella and Badeer, Gada and Swee, Georgia and Halpern, Gil and Herman, Grant and Sizov, Grigory and Guangyi,  and Zhang,  and Lakshminarayanan, Guna and Inan, Hakan and Shojanazeri, Hamid and Zou, Han and Wang, Hannah and Zha, Hanwen and Habeeb, Haroun and Rudolph, Harrison and Suk, Helen and Aspegren, Henry and Goldman, Hunter and Zhan, Hongyuan and Damlaj, Ibrahim and Molybog, Igor and Tufanov, Igor and Leontiadis, Ilias and Veliche, Irina-Elena and Gat, Itai and Weissman, Jake and Geboski, James and Kohli, James and Lam, Janice and Asher, Japhet and Gaya, Jean-Baptiste and Marcus, Jeff and Tang, Jeff and Chan, Jennifer and Zhen, Jenny and Reizenstein, Jeremy and Teboul, Jeremy and Zhong, Jessica and Jin, Jian and Yang, Jingyi and Cummings, Joe and Carvill, Jon and Shepard, Jon and McPhie, Jonathan and Torres, Jonathan and Ginsburg, Josh and Wang, Junjie and Wu, Kai and U, Kam Hou and Saxena, Karan and Khandelwal, Kartikay and Zand, Katayoun and Matosich, Kathy and Veeraraghavan, Kaushik and Michelena, Kelly and Li, Keqian and Jagadeesh, Kiran and Huang, Kun and Chawla, Kunal and Huang, Kyle and Chen, Lailin and Garg, Lakshya and A, Lavender and Silva, Leandro and Bell, Lee and Zhang, Lei and Guo, Liangpeng and Yu, Licheng and Moshkovich, Liron and Wehrstedt, Luca and Khabsa, Madian and Avalani, Manav and Bhatt, Manish and Mankus, Martynas and Hasson, Matan and Lennie, Matthew and Reso, Matthias and Groshev, Maxim and Naumov, Maxim and Lathi, Maya and Keneally, Meghan and Liu, Miao and Seltzer, Michael L. and Valko, Michal and Restrepo, Michelle and Patel, Mihir and Vyatskov, Mik and Samvelyan, Mikayel and Clark, Mike and Macey, Mike and Wang, Mike and Hermoso, Miquel Jubert and Metanat, Mo and Rastegari, Mohammad and Bansal, Munish and Santhanam, Nandhini and Parks, Natascha and White, Natasha and Bawa, Navyata and Singhal, Nayan and Egebo, Nick and Usunier, Nicolas and Mehta, Nikhil and Laptev, Nikolay Pavlovich and Dong, Ning and Cheng, Norman and Chernoguz, Oleg and Hart, Olivia and Salpekar, Omkar and Kalinli, Ozlem and Kent, Parkin and Parekh, Parth and Saab, Paul and Balaji, Pavan and Rittner, Pedro and Bontrager, Philip and Roux, Pierre and Dollar, Piotr and Zvyagina, Polina and Ratanchandani, Prashant and Yuvraj, Pritish and Liang, Qian and Alao, Rachad and Rodriguez, Rachel and Ayub, Rafi and Murthy, Raghotham and Nayani, Raghu and Mitra, Rahul and Parthasarathy, Rangaprabhu and Li, Raymond and Hogan, Rebekkah and Battey, Robin and Wang, Rocky and Howes, Russ and Rinott, Ruty and Mehta, Sachin and Siby, Sachin and Bondu, Sai Jayesh and Datta, Samyak and Chugh, Sara and Hunt, Sara and Dhillon, Sargun and Sidorov, Sasha and Pan, Satadru and Mahajan, Saurabh and Verma, Saurabh and Yamamoto, Seiji and Ramaswamy, Sharadh and Lindsay, Shaun and Lindsay, Shaun and Feng, Sheng and Lin, Shenghao and Zha, Shengxin Cindy and Patil, Shishir and Shankar, Shiva and Zhang, Shuqiang and Zhang, Shuqiang and Wang, Sinong and Agarwal, Sneha and Sajuyigbe, Soji and Chintala, Soumith and Max, Stephanie and Chen, Stephen and Kehoe, Steve and Satterfield, Steve and Govindaprasad, Sudarshan and Gupta, Sumit and Deng, Summer and Cho, Sungmin and Virk, Sunny and Subramanian, Suraj and Choudhury, Sy and Goldman, Sydney and Remez, Tal and Glaser, Tamar and Best, Tamara and Koehler, Thilo and Robinson, Thomas and Li, Tianhe and Zhang, Tianjun and Matthews, Tim and Chou, Timothy and Shaked, Tzook and Vontimitta, Varun and Ajayi, Victoria and Montanez, Victoria and Mohan, Vijai and Kumar, Vinay Satish and Mangla, Vishal and Ionescu, Vlad and Poenaru, Vlad and Mihailescu, Vlad Tiberiu and Ivanov, Vladimir and Li, Wei and Wang, Wenchen and Jiang, Wenwen and Bouaziz, Wes and Constable, Will and Tang, Xiaocheng and Wu, Xiaojian and Wang, Xiaolan and Wu, Xilun and Gao, Xinbo and Kleinman, Yaniv and Chen, Yanjun and Hu, Ye and Jia, Ye and Qi, Ye and Li, Yenda and Zhang, Yilin and Zhang, Ying and Adi, Yossi and Nam, Youngjin and Yu,  and {Wang} and Zhao, Yu and Hao, Yuchen and Qian, Yundi and Li, Yunlu and He, Yuzi and Rait, Zach and DeVito, Zachary and Rosnbrick, Zef and Wen, Zhaoduo and Yang, Zhenyu and Zhao, Zhiwei and Ma, Zhiyu},
  title = {The {Llama} 3 Herd of Models},
  year = {2024},
  howpublished = {Preprint at \url{https://arxiv.org/abs/2407.21783}}
}

@misc{Mistral2023,
  doi = {10.48550/ARXIV.2310.06825},
  author = {Jiang, Albert Q. and Sablayrolles, Alexandre and Mensch, Arthur and Bamford, Chris and Chaplot, Devendra Singh and Casas, Diego de las and Bressand, Florian and Lengyel, Gianna and Lample, Guillaume and Saulnier, Lucile and Lavaud, Lélio Renard and Lachaux, Marie-Anne and Stock, Pierre and Scao, Teven Le and Lavril, Thibaut and Wang, Thomas and Lacroix, Timothée and Sayed, William El},
  title = {{Mistral} 7B},
  year = {2023},
  howpublished = {Preprint at \url{https://arxiv.org/abs/2310.06825}}
}

@misc{Phi3,
  doi = {10.48550/ARXIV.2404.14219},
  author = {Abdin, Marah and Aneja, Jyoti and Awadalla, Hany and Awadallah, Ahmed and Awan, Ammar Ahmad and Bach, Nguyen and Bahree, Amit and Bakhtiari, Arash and Bao, Jianmin and Behl, Harkirat and Benhaim, Alon and Bilenko, Misha and Bjorck, Johan and Bubeck, Sébastien and Cai, Martin and Cai, Qin and Chaudhary, Vishrav and Chen, Dong and Chen, Dongdong and Chen, Weizhu and Chen, Yen-Chun and Chen, Yi-Ling and Cheng, Hao and Chopra, Parul and Dai, Xiyang and Dixon, Matthew and Eldan, Ronen and Fragoso, Victor and Gao, Jianfeng and Gao, Mei and Gao, Min and Garg, Amit and Del Giorno, Allie and Goswami, Abhishek and Gunasekar, Suriya and Haider, Emman and Hao, Junheng and Hewett, Russell J. and Hu, Wenxiang and Huynh, Jamie and Iter, Dan and Jacobs, Sam Ade and Javaheripi, Mojan and Jin, Xin and Karampatziakis, Nikos and Kauffmann, Piero and Khademi, Mahoud and Kim, Dongwoo and Kim, Young Jin and Kurilenko, Lev and Lee, James R. and Lee, Yin Tat and Li, Yuanzhi and Li, Yunsheng and Liang, Chen and Liden, Lars and Lin, Xihui and Lin, Zeqi and Liu, Ce and Liu, Liyuan and Liu, Mengchen and Liu, Weishung and Liu, Xiaodong and Luo, Chong and Madan, Piyush and Mahmoudzadeh, Ali and Majercak, David and Mazzola, Matt and Mendes, Caio César Teodoro and Mitra, Arindam and Modi, Hardik and Nguyen, Anh and Norick, Brandon and Patra, Barun and Perez-Becker, Daniel and Portet, Thomas and Pryzant, Reid and Qin, Heyang and Radmilac, Marko and Ren, Liliang and de Rosa, Gustavo and Rosset, Corby and Roy, Sambudha and Ruwase, Olatunji and Saarikivi, Olli and Saied, Amin and Salim, Adil and Santacroce, Michael and Shah, Shital and Shang, Ning and Sharma, Hiteshi and Shen, Yelong and Shukla, Swadheen and Song, Xia and Tanaka, Masahiro and Tupini, Andrea and Vaddamanu, Praneetha and Wang, Chunyu and Wang, Guanhua and Wang, Lijuan and Wang, Shuohang and Wang, Xin and Wang, Yu and Ward, Rachel and Wen, Wen and Witte, Philipp and Wu, Haiping and Wu, Xiaoxia and Wyatt, Michael and Xiao, Bin and Xu, Can and Xu, Jiahang and Xu, Weijian and Xue, Jilong and Yadav, Sonali and Yang, Fan and Yang, Jianwei and Yang, Yifan and Yang, Ziyi and Yu, Donghan and Yuan, Lu and Zhang, Chenruidong and Zhang, Cyril and Zhang, Jianwen and Zhang, Li Lyna and Zhang, Yi and Zhang, Yue and Zhang, Yunan and Zhou, Xiren},
  title = {{Phi-3} Technical Report: A Highly Capable Language Model Locally on Your Phone},
  year = {2024},
  howpublished = {Preprint at \url{https://arxiv.org/abs/2404.14219}}
}

@article{DeepSeekR1, title={{DeepSeek-R1} incentivizes reasoning in {LLMs} through reinforcement learning}, volume={645}, ISSN={1476-4687}, DOI={10.1038/s41586-025-09422-z}, number={8081}, journal={Nature}, publisher={Springer Science and Business Media LLC}, author={Guo, Daya and Yang, Dejian and Zhang, Haowei and Song, Junxiao and Wang, Peiyi and Zhu, Qihao and Xu, Runxin and Zhang, Ruoyu and Ma, Shirong and Bi, Xiao and Zhang, Xiaokang and Yu, Xingkai and Wu, Yu and Wu, Z. F. and Gou, Zhibin and Shao, Zhihong and Li, Zhuoshu and Gao, Ziyi and Liu, Aixin and Xue, Bing and Wang, Bingxuan and Wu, Bochao and Feng, Bei and Lu, Chengda and Zhao, Chenggang and Deng, Chengqi and Ruan, Chong and Dai, Damai and Chen, Deli and Ji, Dongjie and Li, Erhang and Lin, Fangyun and Dai, Fucong and Luo, Fuli and Hao, Guangbo and Chen, Guanting and Li, Guowei and Zhang, H. and Xu, Hanwei and Ding, Honghui and Gao, Huazuo and Qu, Hui and Li, Hui and Guo, Jianzhong and Li, Jiashi and Chen, Jingchang and Yuan, Jingyang and Tu, Jinhao and Qiu, Junjie and Li, Junlong and Cai, J. L. and Ni, Jiaqi and Liang, Jian and Chen, Jin and Dong, Kai and Hu, Kai and You, Kaichao and Gao, Kaige and Guan, Kang and Huang, Kexin and Yu, Kuai and Wang, Lean and Zhang, Lecong and Zhao, Liang and Wang, Litong and Zhang, Liyue and Xu, Lei and Xia, Leyi and Zhang, Mingchuan and Zhang, Minghua and Tang, Minghui and Zhou, Mingxu and Li, Meng and Wang, Miaojun and Li, Mingming and Tian, Ning and Huang, Panpan and Zhang, Peng and Wang, Qiancheng and Chen, Qinyu and Du, Qiushi and Ge, Ruiqi and Zhang, Ruisong and Pan, Ruizhe and Wang, Runji and Chen, R. J. and Jin, R. L. and Chen, Ruyi and Lu, Shanghao and Zhou, Shangyan and Chen, Shanhuang and Ye, Shengfeng and Wang, Shiyu and Yu, Shuiping and Zhou, Shunfeng and Pan, Shuting and Li, S. S. and Zhou, Shuang and Wu, Shaoqing and Yun, Tao and Pei, Tian and Sun, Tianyu and Wang, T. and Zeng, Wangding and Liu, Wen and Liang, Wenfeng and Gao, Wenjun and Yu, Wenqin and Zhang, Wentao and Xiao, W. L. and An, Wei and Liu, Xiaodong and Wang, Xiaohan and Chen, Xiaokang and Nie, Xiaotao and Cheng, Xin and Liu, Xin and Xie, Xin and Liu, Xingchao and Yang, Xinyu and Li, Xinyuan and Su, Xuecheng and Lin, Xuheng and Li, X. Q. and Jin, Xiangyue and Shen, Xiaojin and Chen, Xiaosha and Sun, Xiaowen and Wang, Xiaoxiang and Song, Xinnan and Zhou, Xinyi and Wang, Xianzu and Shan, Xinxia and Li, Y. K. and Wang, Y. Q. and Wei, Y. X. and Zhang, Yang and Xu, Yanhong and Li, Yao and Zhao, Yao and Sun, Yaofeng and Wang, Yaohui and Yu, Yi and Zhang, Yichao and Shi, Yifan and Xiong, Yiliang and He, Ying and Piao, Yishi and Wang, Yisong and Tan, Yixuan and Ma, Yiyang and Liu, Yiyuan and Guo, Yongqiang and Ou, Yuan and Wang, Yuduan and Gong, Yue and Zou, Yuheng and He, Yujia and Xiong, Yunfan and Luo, Yuxiang and You, Yuxiang and Liu, Yuxuan and Zhou, Yuyang and Zhu, Y. X. and Huang, Yanping and Li, Yaohui and Zheng, Yi and Zhu, Yuchen and Ma, Yunxian and Tang, Ying and Zha, Yukun and Yan, Yuting and Ren, Z. Z. and Ren, Zehui and Sha, Zhangli and Fu, Zhe and Xu, Zhean and Xie, Zhenda and Zhang, Zhengyan and Hao, Zhewen and Ma, Zhicheng and Yan, Zhigang and Wu, Zhiyu and Gu, Zihui and Zhu, Zijia and Liu, Zijun and Li, Zilin and Xie, Ziwei and Song, Ziyang and Pan, Zizheng and Huang, Zhen and Xu, Zhipeng and Zhang, Zhongyu and Zhang, Zhen}, year={2025}, pages={633–638},
  url = {https://doi.org/10.1038/s41586-025-09422-z}
}

@misc{GptOss2025,
  doi = {10.48550/ARXIV.2508.10925},
  author = {{OpenAI} and Agarwal, Sandhini and Ahmad, Lama and Ai, Jason and Altman, Sam and Applebaum, Andy and Arbus, Edwin and Arora, Rahul K. and Bai, Yu and Baker, Bowen and Bao, Haiming and Barak, Boaz and Bennett, Ally and Bertao, Tyler and Brett, Nivedita and Brevdo, Eugene and Brockman, Greg and Bubeck, Sebastien and Chang, Che and Chen, Kai and Chen, Mark and Cheung, Enoch and Clark, Aidan and Cook, Dan and Dukhan, Marat and Dvorak, Casey and Fives, Kevin and Fomenko, Vlad and Garipov, Timur and Georgiev, Kristian and Glaese, Mia and Gogineni, Tarun and Goucher, Adam and Gross, Lukas and Guzman, Katia Gil and Hallman, John and Hehir, Jackie and Heidecke, Johannes and Helyar, Alec and Hu, Haitang and Huet, Romain and Huh, Jacob and Jain, Saachi and Johnson, Zach and Koch, Chris and Kofman, Irina and Kundel, Dominik and Kwon, Jason and Kyrylov, Volodymyr and Le, Elaine Ya and Leclerc, Guillaume and Lennon, James Park and Lessans, Scott and Lezcano-Casado, Mario and Li, Yuanzhi and Li, Zhuohan and Lin, Ji and Liss, Jordan and Liu, Lily (Xiaoxuan) and Liu, Jiancheng and Lu, Kevin and Lu, Chris and Martinovic, Zoran and McCallum, Lindsay and McGrath, Josh and McKinney, Scott and McLaughlin, Aidan and Mei, Song and Mostovoy, Steve and Mu, Tong and Myles, Gideon and Neitz, Alexander and Nichol, Alex and Pachocki, Jakub and Paino, Alex and Palmie, Dana and Pantuliano, Ashley and Parascandolo, Giambattista and Park, Jongsoo and Pathak, Leher and Paz, Carolina and Peran, Ludovic and Pimenov, Dmitry and Pokrass, Michelle and Proehl, Elizabeth and Qiu, Huida and Raila, Gaby and Raso, Filippo and Ren, Hongyu and Richardson, Kimmy and Robinson, David and Rotsted, Bob and Salman, Hadi and Sanjeev, Suvansh and Schwarzer, Max and Sculley, D. and Sikchi, Harshit and Simon, Kendal and Singhal, Karan and Song, Yang and Stuckey, Dane and Sun, Zhiqing and Tillet, Philippe and Toizer, Sam and Tsimpourlas, Foivos and Vyas, Nikhil and Wallace, Eric and Wang, Xin and Wang, Miles and Watkins, Olivia and Weil, Kevin and Wendling, Amy and Whinnery, Kevin and Whitney, Cedric and Wong, Hannah and Yang, Lin and Yang, Yu and Yasunaga, Michihiro and Ying, Kristen and Zaremba, Wojciech and Zhan, Wenting and Zhang, Cyril and Zhang, Brian and Zhang, Eddie and Zhao, Shengjia},
  title = {gpt-oss-120b \& gpt-oss-20b Model Card},
  year = {2025},
  howpublished = {Preprint at \url{https://arxiv.org/abs/2508.10925}}
}

@inproceedings{Wolf2020, title={{Transformers}: State-of-the-Art Natural Language Processing}, DOI={10.18653/v1/2020.emnlp-demos.6}, booktitle={Proceedings of the 2020 Conference on Empirical Methods in Natural Language Processing: System Demonstrations}, publisher={Association for Computational Linguistics}, author={Wolf, Thomas and Debut, Lysandre and Sanh, Victor and Chaumond, Julien and Delangue, Clement and Moi, Anthony and Cistac, Pierric and Rault, Tim and Louf, Remi and Funtowicz, Morgan and Davison, Joe and Shleifer, Sam and von Platen, Patrick and Ma, Clara and Jernite, Yacine and Plu, Julien and Xu, Canwen and Le Scao, Teven and Gugger, Sylvain and Drame, Mariama and Lhoest, Quentin and Rush, Alexander}, year={2020}, pages={38–45},
  url = {https://doi.org/10.18653/v1/2020.emnlp-demos.6}
}

@incollection{Lin2023alliance,
 author = {Lin, Baihan and Cecchi, Guillermo and Bouneffouf, Djallel},
 title = {Deep annotation of therapeutic working alliance in psychotherapy},
 booktitle = {Artificial Intelligence for Personalized Medicine: Promoting Healthy Living and Longevity},
 editor = {Shaban-Nejad, Arash and Michalowski, Martin and Bianco, Simone},
 series = {Studies in Computational Intelligence}, volume = {1106},
 publisher = {Springer}, address = {Cham}, year = {2023}, pages = {193--207},
 doi = {10.1007/978-3-031-36938-4_15}}

@incollection{Lin2023topics,
 author = {Lin, Baihan and Bouneffouf, Djallel and Cecchi, Guillermo and Tejwani, Ravi},
 title = {Neural topic modeling of psychotherapy sessions},
 booktitle = {Artificial Intelligence for Personalized Medicine: Promoting Healthy Living and Longevity},
 editor = {Shaban-Nejad, Arash and Michalowski, Martin and Bianco, Simone},
 series = {Studies in Computational Intelligence}, volume = {1106},
 publisher = {Springer}, address = {Cham}, year = {2023}, pages = {209--219},
 doi = {10.1007/978-3-031-36938-4_16}}

@inproceedings{Lin2024wat,
 author = {Lin, Baihan and Cecchi, Guillermo and Bouneffouf, Djallel},
 title = {Working alliance transformer for psychotherapy dialogue classification},
 booktitle = {Proceedings of the 6th Clinical Natural Language Processing Workshop},
 publisher = {Association for Computational Linguistics}, address = {Mexico City},
 year = {2024}, pages = {64--69}, doi = {10.18653/v1/2024.clinicalnlp-1.6}}

@inproceedings{Gunal2024decision,
 author = {Gunal, Aylin and Lin, Baihan and Bouneffouf, Djallel},
 title = {Conversational topic recommendation in counseling and psychotherapy with decision transformer and large language models},
 booktitle = {Proceedings of the 6th Clinical Natural Language Processing Workshop},
 publisher = {Association for Computational Linguistics}, address = {Mexico City},
 year = {2024}, pages = {196--201}, doi = {10.18653/v1/2024.clinicalnlp-1.16}}

@inproceedings{Lin2023supervisor,
 author = {Lin, Baihan and Cecchi, Guillermo and Bouneffouf, Djallel},
 title = {{SupervisorBot}: {NLP}-annotated real-time recommendations of psychotherapy treatment strategies with deep reinforcement learning},
 booktitle = {Proceedings of the Thirty-Second International Joint Conference on Artificial Intelligence},
 publisher = {International Joint Conferences on Artificial Intelligence Organization},
 year = {2023}, pages = {7149--7153}, doi = {10.24963/ijcai.2023/837}}

@inproceedings{Lin2023companion,
 author = {Lin, Baihan and Cecchi, Guillermo and Bouneffouf, Djallel},
 title = {Psychotherapy {AI} companion with reinforcement learning recommendations and interpretable policy dynamics},
 booktitle = {Companion Proceedings of the ACM Web Conference 2023},
 publisher = {Association for Computing Machinery}, address = {New York},
 year = {2023}, pages = {932--939}, doi = {10.1145/3543873.3587623}}

@misc{Lin2026compass2,
 author = {Lin, Baihan},
 title = {{COMPASS} 2.0: psychometric representational similarity analysis distinguishes symptom structure from personal signal},
 year = {2026}, howpublished = {Preprint at https://arxiv.org/abs/2610.06615},
 url = {https://arxiv.org/abs/2610.06615}}

@inproceedings{Gratch2014,
  author = {Gratch, Jonathan and Artstein, Ron and Lucas, Gale and Stratou, Giota and Scherer, Stefan and Nazarian, Angela and Wood, Rachel and Boberg, Jill and DeVault, David and Marsella, Stacy and Traum, David and Rizzo, Skip and Morency, Louis-Philippe},
  title = {The Distress Analysis Interview Corpus of human and computer interviews},
  booktitle = {Proceedings of the Ninth International Conference on Language Resources and Evaluation (LREC'14)},
  pages = {3123--3128}, year = {2014},
  url = {https://aclanthology.org/L14-1421/}
}

@inproceedings{DeVault2014,
  author = {DeVault, David and Artstein, Ron and Benn, Grace and Dey, Teresa and Fast, Ed and Gainer, Alesia and Georgila, Kallirroi and Gratch, Jon and Hartholt, Arno and Lhommet, Margaux and Lucas, Gale and Marsella, Stacy and Morbini, Fabrizio and Nazarian, Angela and Scherer, Stefan and Stratou, Giota and Suri, Apar and Traum, David and Wood, Rachel and Xu, Yuyu and Rizzo, Albert and Morency, Louis-Philippe},
  title = {{SimSensei} {Kiosk}: a virtual human interviewer for healthcare decision support},
  booktitle = {Proceedings of the 2014 International Conference on Autonomous Agents and Multi-Agent Systems (AAMAS)},
  pages = {1061--1068}, year = {2014},
  url = {https://aamas.csc.liv.ac.uk/Proceedings/aamas2014/aamas/p1061.pdf}
}

@inproceedings{LechugaLopez2026,
  author = {Lechuga L\'{o}pez, L. Juli\'{a}n and Shamout, Farah E. and Rudner, Tim G. J.},
  title = {An empirical analysis of calibration and selective prediction in multimodal clinical condition classification},
  booktitle = {Proceedings of the 7th Conference on Health, Inference, and Learning},
  series = {Proceedings of Machine Learning Research}, volume = {333}, pages = {794--833}, year = {2026},
  url = {https://proceedings.mlr.press/v333/lopez26a.html}
}

@book{Cronbach1972,
  author = {Cronbach, Lee J. and Gleser, Goldine C. and Nanda, Harinder and Rajaratnam, Nageswari},
  title = {The Dependability of Behavioral Measurements: Theory of Generalizability for Scores and Profiles},
  publisher = {Wiley}, address = {New York}, year = {1972},
  url = {https://gwern.net/doc/psychology/1972-cronbach-thedependabilityofbehavioralmeasurements.pdf}
}

\end{document}